\pdfoutput=1
\documentclass[lettersize,journal]{IEEEtran}

\usepackage{colortbl}
\usepackage{bm}
\usepackage{amsmath}
\usepackage{wrapfig}
\usepackage[T1]{fontenc}
\usepackage{soul}
\usepackage{multirow}
\usepackage{makecell}
\usepackage{booktabs}
\usepackage{nicefrac}
\usepackage{wrapfig}
\usepackage[noend]{algpseudocode}
\usepackage[ruled]{algorithm2e}
\usepackage[scaled=0.95]{inconsolata}
\usepackage{amssymb}
\usepackage{etoolbox}

\usepackage{enumitem}
\usepackage{graphicx}
\usepackage[font={small}]{caption}
\usepackage[dvipsnames]{xcolor} 
\usepackage{hyperref}
\hypersetup{
    colorlinks=true,
    citecolor=LimeGreen,
    linkcolor=WildStrawberry,
    urlcolor=cyan,
}
\usepackage{listings}
\usepackage{subcaption}
\usepackage{tabularx}
\usepackage{tikz}
\usetikzlibrary{fadings}
\usepackage{titlesec}
\usepackage{soul}
\usepackage{multirow}
\usepackage{wrapfig}

\usepackage{amsfonts,bm}

\def\eqref#1{equation~\ref{#1}}

\def\1{\bm{1}}

\DeclareMathAlphabet{\mathsfit}{\encodingdefault}{\sfdefault}{m}{sl}
\SetMathAlphabet{\mathsfit}{bold}{\encodingdefault}{\sfdefault}{bx}{n}

\DeclareMathOperator*{\argmax}{arg\,max}

\definecolor{deepblue}{rgb}{0,0,0.5}
\definecolor{deepred}{rgb}{0.6,0,0}
\definecolor{magenta}{rgb}{1.0,0,1.0}
\definecolor{deepgreen}{rgb}{0,0.5,0}
\definecolor{textblue}{rgb}{.2,.2,.7}
\definecolor{textred}{rgb}{0.54,0,0}
\definecolor{textgreen}{rgb}{0,0.43,0}
\definecolor{es-blue}{rgb}{0.1372,0.666,1}
\definecolor{allcolor}{rgb}{1.0,0.53,0.0}
\definecolor{author}{rgb}{0.4,0.3,0.7}
\definecolor{bestgreen}{RGB}{180,230,180}
\definecolor{secondgreen}{RGB}{225,245,225}
\definecolor{tablecolor}{RGB}{189,173,248}
\definecolor{microwavecolor}{RGB}{180,229,162}
\definecolor{kitchencolor}{RGB}{243,185,150}
\definecolor{projectlink}{rgb}{0.90,0.55,0.75}
\newcolumntype{Y}{>{\centering\arraybackslash}X}
\newcommand{\resulttablewidth}{0.85\columnwidth}

\usepackage[capitalise, nameinlink]{cleveref}
\makeatletter
\let\NAT@parse\undefined
\makeatother
\usepackage{xspace}

\newcommand{\authorhref}[3][author]{\href{#2}{\color{#1}{#3}}}

\newcommand{\best}[1]{\cellcolor{bestgreen}\textbf{#1}}
\newcommand{\second}[1]{\cellcolor{secondgreen}#1}

\usepackage{cite}
\usepackage{stfloats}
\providecommand{\acknowledgments}{}
\renewcommand{\acknowledgments}[1]{\section*{Acknowledgments}#1}

\makeatletter
\renewcommand\paragraph{\@startsection{paragraph}{4}{\z@}{1.5ex plus 1.5ex minus 0.5ex}{-0.5em}{\normalfont\normalsize\bfseries}}
\def\@IEEEsectpunct{\ }
\makeatother
\newcommand{\videourl}{https://ethz-mrl.github.io/robot-self-improvement-website/}

\begin{document}

\title{\textit{Learning from Failure:} \texorpdfstring{\\}{ } Robot Self-Improvement via Human Videos}

\author{
\textbf{\authorhref{https://hanzhic.github.io/}{Hanzhi Chen}}$^{\star,1,2,4}$ \quad
\textbf{\authorhref{https://dipan-zhang.github.io/}{Anran Zhang}}$^{\star,1,2,4}$ \quad
\textbf{\authorhref{https://simon-schaefer.github.io/}{Simon Schaefer}}$^{1,2,4}$ \quad
\textbf{\authorhref{https://www.ce.cit.tum.de/air/people/kejia-chen-msc/}{Kejia Chen}}$^{2}$ \\
\textbf{\authorhref{https://scholar.google.com/citations?user=rAO6AHsAAAAJ}{Shi Chen}}$^{1}$ \quad
\textbf{\authorhref{https://scholar.google.com/citations?user=cXQciMEAAAAJ}{Daniel Cremers}}$^{2,4}$ \quad
\textbf{\authorhref{https://www.oiermees.com/}{Oier Mees}}$^{\dagger,3,1}$ \quad
\textbf{\authorhref{https://scholar.google.ch/citations?user=SmGQ48gAAAAJ}{Stefan Leutenegger}}$^{\dagger,1}$ \\[5pt]
{\normalfont $^{\star}$Equal contribution \quad $^{\dagger}$Equal advising \quad Order within each group is interchangeable.} \\[5pt]
{\normalfont $^{1}$ETH Zurich \quad
$^{2}$Technical University of Munich \quad
$^{3}$Microsoft \quad
$^{4}$MCML} \\[5pt]
{\ttfamily \href{\videourl}{\textcolor{projectlink}{\nolinkurl{\videourl}}}}
}

\markboth{}%
{Chen \MakeLowercase{\textit{et al.}}: Learning from Failure: Robot Self-Improvement via Human Videos}

\IEEEaftertitletext{\vspace{-2\baselineskip}}
\maketitle

\begin{abstract}
Learning from deployment experience is essential for robots to improve their performance in the real world. A robot should be able to learn from failed executions, but these do not reveal which actions would have succeeded, and obtaining corrections from human experts requires costly monitoring and intervention. Human videos offer a rich source of interaction experience, showing how actions change the environment and lead to task completion. To turn this experience into corrective supervision, we introduce Dynamics-Guided Action Correction (DGAC), which uses dynamics and value models pretrained on human videos to construct alternative actions at failed states and evaluate their outcomes without executing each candidate. We evaluate our method on seven real-world manipulation tasks across two robot embodiments and demonstrate that it improves the average success rate from 40\% to 81\%, requiring no expert corrections during self-improvement. We further show that generated corrections enable recovery from failed executions and provide more effective training targets than imagined policy rollouts using the same models and data. Finally, we demonstrate that human-video dynamics and value models provide effective corrective supervision for policies pretrained on either human or robot data, suggesting that human videos benefit robot learning in addition to policy initialization.

\end{abstract}


\section{Introduction}
\label{sec:introduction}

Robots should improve from their own failures. Policies learned from a limited set of demonstrations often fail when execution departs from the demonstrated behavior, and these failures reveal where the policy must change. A failed rollout, however, records what the robot did, not which alternative actions would have succeeded. Human-in-the-loop methods supply corrective actions~\cite{kelly2019hg,jang2022bc,wu2025robocopilot,luo2025precise,intelligence2025pi}, but require experts to monitor executions and provide corrective interventions. Maintaining this level of expert oversight becomes increasingly difficult as robots are deployed across more tasks and environments. This raises a central question for robot self-improvement: \emph{how can a robot construct corrective action supervision from its own failed rollouts?}

Human videos offer interaction experience at scale that could help supply this missing supervision: they show how actions change a scene and how tasks progress across diverse objects and environments. Prior work has used this experience to learn visual representations~\cite{nair2022r3m,xiao2022masked,ma2022vip}, actionable interaction structures for robot policies~\cite{chen2025vidbot,xu2024flow,zhang2025actron3d,yang2025egovla}, and world models for planning~\cite{mendonca2023SWIM}. We investigate a complementary use of human videos: learning models that help determine what a robot should have done differently when its policy fails. A dynamics model can predict the consequences of alternative actions from a recorded state, while a value model evaluates whether those consequences lead toward task success. Together, these models allow candidate corrections to be compared without repeatedly restoring the failed state and executing each alternative. Learning them from human videos could reduce reliance on expert corrective actions. 

\begin{figure}[t]
\centering
\includegraphics[width=\linewidth]{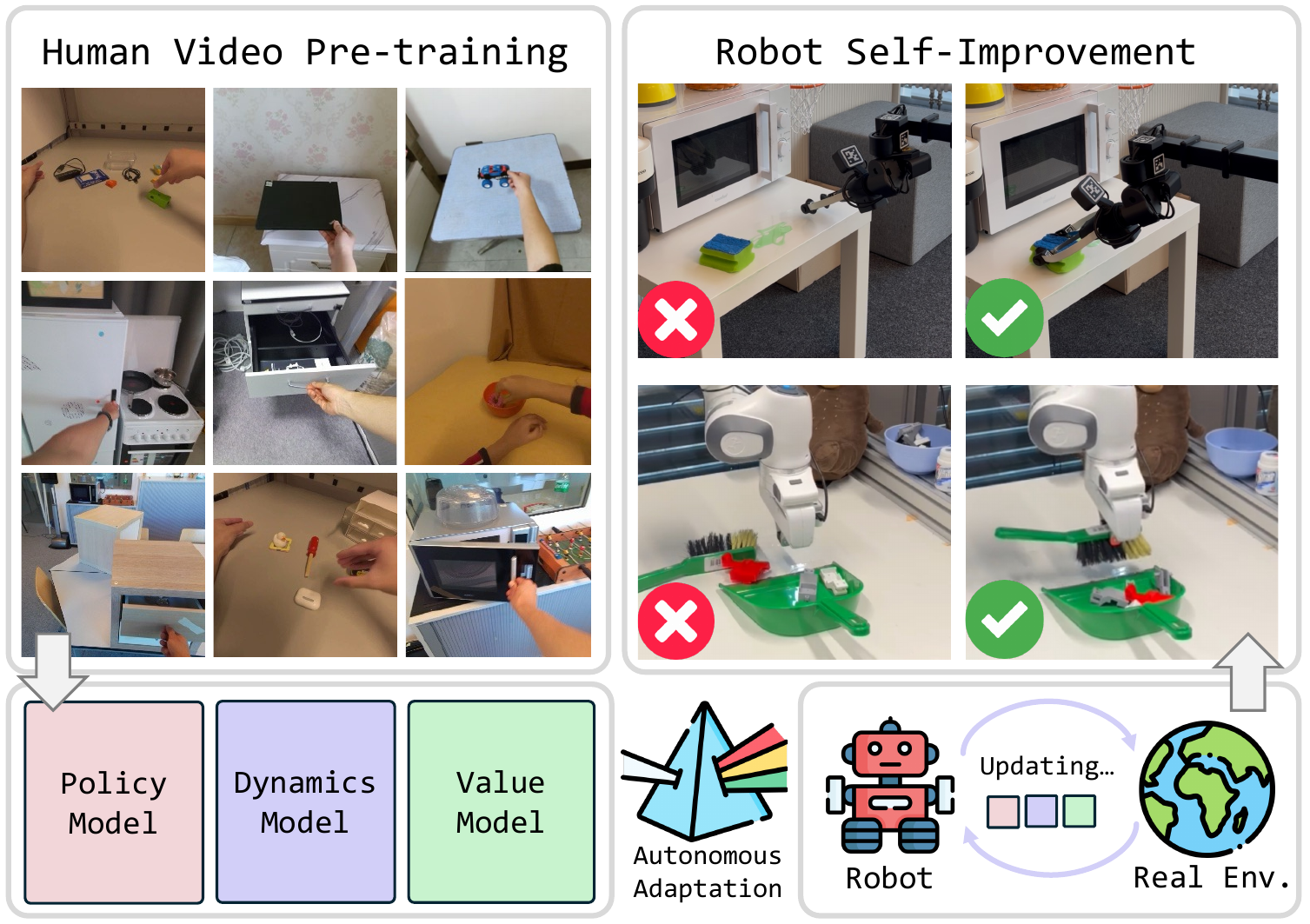}
\caption{Policy, dynamics, and value models are pretrained on human videos (left), adapted with robot experience, and used in an iterative loop that turns failed rollouts into corrective supervision for policy updates. The examples show failed and successful executions on the Stretch (top) and Franka (bottom) robots.}
\label{fig:teaser}
\end{figure}

Predicting action outcomes is only part of the problem: the robot must also propose actions that could correct the failure. Sampling from the current policy may reproduce its mistakes, while copying an action from a successful execution may not suit the current scene. To address this challenge, we introduce \emph{Dynamics-Guided Action Correction} (DGAC). Given a state from a failed rollout, DGAC retrieves a similar state from successful robot experience and uses it to guide action generation conditioned on the current observation. The dynamics model predicts each candidate's outcome and the value model selects the most promising candidate as a corrective target. 

We pretrain policy, dynamics, and value models on human videos using representations shared by humans and robots, then adapt them with robot experience (Fig.~\ref{fig:teaser}). This adaptation addresses differences in embodiment and exposes the models to robot failures that are underrepresented in human videos. Policy improvement then alternates between collecting rollouts, updating the dynamics and value models, generating corrections with DGAC, and training the policy on successful experience and corrective state--action examples.

We evaluate the framework on seven real-world manipulation tasks across two embodiments: a mobile manipulator (Hello Robot Stretch 3) and a fixed-base robot arm (Franka Panda). Across these tasks, it reaches 81.0\% average success, compared with 40.0\% for behavior cloning on the same demonstrations. Direct execution of the generated corrections shows that selected actions enable recovery from failures. On the five Stretch tasks, a matched comparison holds the models and rollout data fixed: DGAC provides more effective policy-training supervision than a RISE-style pipeline~\cite{yang2026rise} that synthesizes and value-filters policy rollouts (85.3\% vs.\ 70.7\%).

In summary, our key contributions are:
(1) a self-improvement framework that uses human-video priors, adapted with robot experience, to construct corrective supervision from failed robot
  rollouts;
(2) DGAC, which uses successful experience to guide action proposals at failed states and selects corrective targets by evaluating their predicted
  consequences; and
(3) real-world experiments on two robot embodiments demonstrating that the corrections enable recovery from failures, yield greater policy improvement than alternative supervision generated with matched models and rollout data, and benefit policies pretrained on either human or robot data.

\section{Related Work}
\label{sec:related_work}

\paragraph{Real-World Policy Improvement.}
Human-in-the-loop methods improve robot policies with expert interventions that provide corrective behaviors during execution~\cite{jang2022bc,wu2025robocopilot,kelly2019hg,luo2025precise}.
Reinforcement learning also enables policies to improve through interaction and reward feedback~\cite{kalashnikov2021mt,li2025simplevla,lei2025rl,guo2025improving}.
Residual RL learns corrections to a base policy, either to refine its execution~\cite{ankile2024resip} or to collect improved trajectories for subsequent policy training~\cite{xiao2025self}.
Advantage-weighted and advantage-conditioned policy extraction favor higher-return behavior in the collected data~\cite{peng2019AWR,frans2025diffusion,intelligence2025pi}.
These approaches can also incorporate human corrections: RECAP~\cite{intelligence2025pi} combines demonstrations, autonomous rollouts, and expert interventions through advantage-conditioned policy updates.
Our framework builds on advantage-conditioned policy updates and focuses on generating corrective action supervision from recorded failures.

\paragraph{Robot Learning with Dynamics Models.}
Learned dynamics models support sample-efficient policy optimization~\cite{wu2023daydreamer,nagabandi2020deep,chua2018deep,hafner2019dream,hafner2019learning,nematoli20iros,chandra2025diwa,hou2026world, li2026egowam}. They also enable policy evaluation~\cite{guo2025ctrl,team2025evaluating}, policy steering~\cite{sun2025lpb,qi2026inference,du2025dynaguide}, and synthetic data generation~\cite{jang2025dreamgen}.
RISE~\cite{yang2026rise} uses a compositional world model to generate imagined policy rollouts, estimate their advantages, and update the policy.
CCIL~\cite{ke2023ccil} uses locally continuous dynamics learned from demonstrations to generate corrective labels at nearby states.
DGAC uses learned dynamics to construct corrective supervision from the robot's own failed rollouts.
Learning dynamics across diverse states requires rich interaction data, motivating our use of human-video pretraining followed by adaptation with robot experience.

\paragraph{Learning from Human Videos.}
Human videos offer a promising resource for scaling robot learning, as they are ubiquitous and contain rich priors about object interactions, task progress, and contact dynamics.
Prior works have explored several forms of supervision from human videos~\cite{feng2026human}, including visual representations~\cite{ma2022vip,nair2022r3m,mees20icra_asn}, reward functions~\cite{chen2021learning,shi2025points2reward}, and transferable action representations derived from affordances, hand poses, and motion~\cite{bahl2023affordances,bharadhwaj2024towards,chen2025vidbot,yang2025egovla,zhang2025actron3d,xu2024flow}.
WHIRL~\cite{bahl2022human} uses human demonstrations to guide policy improvement through real-world interaction.
Human videos also support dynamics learning: SWIM~\cite{mendonca2023SWIM} pretrains a world model on human videos, adapts it with robot interaction data, and uses it for planning.
We investigate whether human-video priors can help policies improve from their own failures by supporting the construction of corrective supervision.

\section{Preliminaries}
\label{sec:preliminaries}

\paragraph{Flow Matching.}
Flow matching learns a time-dependent velocity field that transports a noise distribution toward the data distribution~\cite{lipman2022flow}. Given clean data $\bm{x}$ and Gaussian noise $\bm{\epsilon}\sim\mathcal{N}(\mathbf{0},\mathbf{I})$, a noisy sample at timestamp $\tau\in[0,1]$ is defined by linear interpolation $\bm{x}^{\tau}=\tau\bm{x}+(1-\tau)\bm{\epsilon}$, with the corresponding velocity $\bm{v}=\frac{d\bm{x}^{\tau}}{d\tau}=\bm{x}-\bm{\epsilon}$. The model $\bm{v}_\theta(\bm{x}^{\tau},\tau)$ is trained to match this velocity field, and generation is performed by integrating the resulting ODE from noise to data.
Following recent work on prediction space~\cite{li2511back}, we adopt the \emph{x-prediction} parameterization: the network directly predicts the clean target $\hat{\bm{x}}_\theta(\bm{x}^\tau,\tau)$, from which the velocity field is recovered as $\bm{v}_\theta(\bm{x}^\tau,\tau)=\nicefrac{\big(\hat{\bm{x}}_\theta(\bm{x}^\tau,\tau)-\bm{x}^\tau\big)}{(1-\tau)}$ for integration.

\paragraph{Reinforcement Learning.}
We formulate the robot manipulation task as a finite-horizon contextual Markov decision process (MDP) with observation space $\mathcal{O}$, action space $\mathcal{A}$, reward function $r$, and episode horizon $T$. At timestep $t$, the policy predicts an action chunk~\cite{zhao2023learning} $\bm{a}_{t:t+H-1}\in\mathcal{A}^H$ conditioned on the current observation $\bm{o}_t\in\mathcal{O}$, thereby inducing a trajectory distribution $\rho_\pi(\tau)$. The RL objective is to maximize the cumulative reward
\(
\mathcal{J}(\pi)=\mathbb{E}_{\tau\sim\rho_\pi}[\sum_{t=0}^{T} r_t].
\)
To assess progress toward successful task completion, we learn a parametric progress value model $V_\psi(\bm{o}_t)$. For advantage-conditioned policy extraction, we score each action chunk with a \emph{progress-based advantage surrogate} that combines the rewards within the chunk with the change in predicted progress value:
\begin{equation}
\label{eqn:advantage}
\hat{A}(\bm{o}_t,\bm{a}_{t:t+H-1})
=
\sum_{t'=t}^{t+H-1} r_{t'}
+
V_\psi(\bm{o}_{t+H})
-
V_\psi(\bm{o}_t).
\end{equation}
We compute this surrogate without additional temporal discounting.
We adopt the policy-extraction view and construct a target policy from a reference policy $\pi_{\mathrm{ref}}$:
\(
\hat{\pi}(\bm{a}_{t:t+H-1}\mid\bm{o}_t)
\propto \pi_{\mathrm{ref}}(\bm{a}_{t:t+H-1}\mid\bm{o}_t)\ p(I_t=1\mid\hat{A}_t),
\)
where $\hat{A}_t=\hat{A}(\bm{o}_t,\bm{a}_{t:t+H-1})$ and $I_t=1$ denotes the improvement event. For training, we assign each sample a binary advantage label $I_t=\mathbf{1}(\hat{A}_t>\kappa)$ using a task-specific threshold $\kappa$, following CFGRL and RECAP~\cite{frans2025diffusion, intelligence2025pi}.

\section{Method}
\label{sec:method}

\begin{figure*}[t]
\centering
\includegraphics[width=\linewidth]{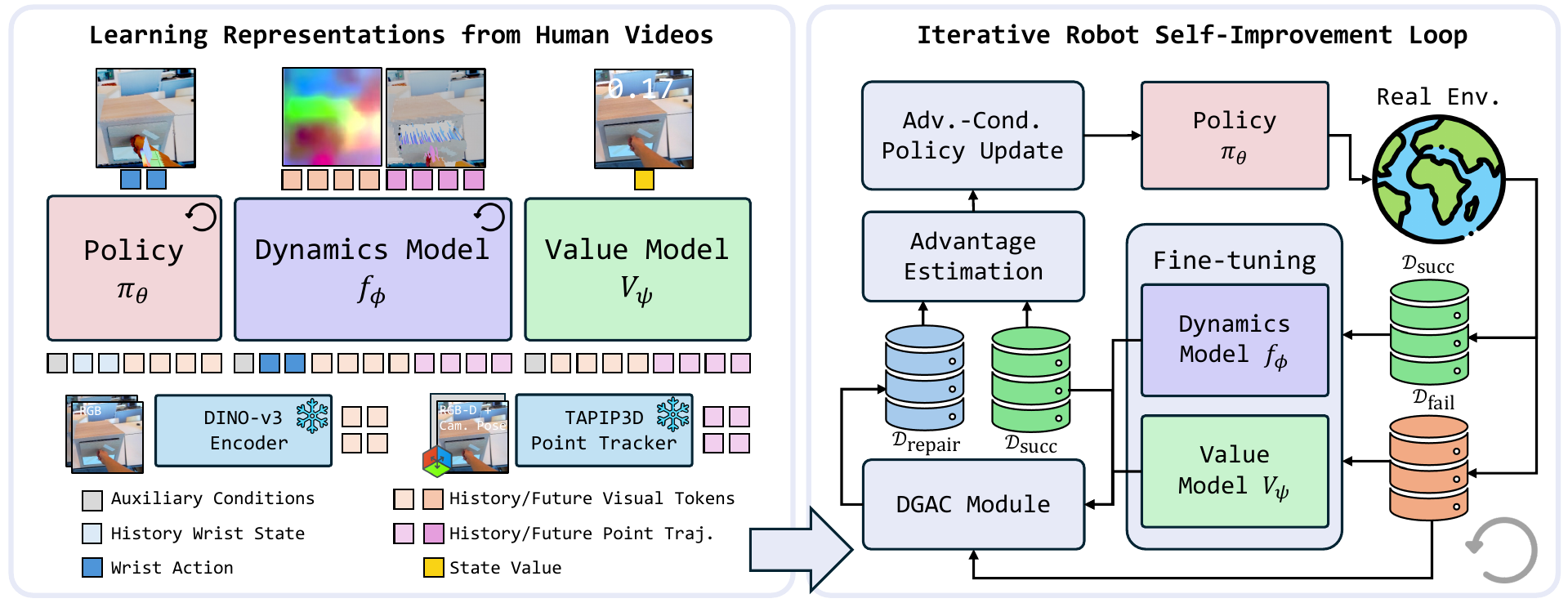}
\caption{
Overview of our framework. We first pretrain reusable policy, dynamics, and value models from human videos, which then enable iterative robot policy self-improvement in the real world.
}
\label{fig:pipeline}
\end{figure*}

Our framework improves a robot policy by constructing corrective supervision from its own failed rollouts (Fig.~\ref{fig:pipeline}). A policy model proposes alternative actions, a dynamics model predicts their outcomes, and a progress value model evaluates progress toward task success. We pretrain these models on human videos and adapt them with robot experience. \textit{Dynamics-Guided Action Correction} (DGAC) uses successful references to guide action generation at failed states and selects corrections using the predicted outcomes. The selected actions become supervision for subsequent policy updates, forming an iterative self-improvement loop.

\subsection{Reusable Structures from Human Videos}
\label{sec:method_models}
We aim to extract reusable structures from human videos for robot learning so that the learned representations transfer across embodiments while remaining amenable to downstream policy improvement. 
We therefore define transferable action, world-state, and value representations, and jointly learn policy, dynamics, and value models for robot self-improvement.
\paragraph{Human--Robot Shared Representations.}
These representations preserve interaction structure while abstracting away embodiment-specific details, enabling prediction, evaluation, and correction across embodiments.
Each action is represented as $\bm{a}_t=[\bm{\xi}_t,c_t]$, where $\bm{\xi}_t\in SE(3)$ denotes the absolute 6-DoF wrist pose and $c_t\in[0,1]$ is a scalar hand-closure variable estimated from human finger configurations. This abstracts away embodiment-specific articulation while retaining key motion cues.

The world state is represented as $\bm{o}_t=[\bm{z}_t,\bm{P}_t]$, where $\bm{z}_t\in\mathbb{R}^{L_z\times D}$ denotes current-frame DINO-v3 visual tokens~\cite{simeoni2025dinov3}, and $\bm{P}_t\in\mathbb{R}^{L_p\times H'\times 3}$ denotes short-horizon 3D point flows~\cite{tapip3d} ending at time $t$. This asymmetric temporal design is compact: semantic
context changes slowly over short horizons, while point flows capture the fine-grained geometry and contact dynamics needed for prediction.  

We train the value function using the discounted return induced by a sparse terminal reward, $V_t = \sum_{t'=t}^{T} \gamma^{t'-t} r_{t'} = \gamma^{T-t} r_T$, where $\gamma$ is the discount factor, $T$ is the terminal timestep, all
intermediate rewards are $0$, and the terminal reward $r_T\in\{1,-1\}$ denotes success or failure.

\paragraph{Policy, Dynamics, and Value Modeling.}
Building on the representations above, we jointly learn a policy model,
a dynamics model, and a value model. We use $H'$ to denote the history
horizon and $H$ to denote the future prediction horizon. For compactness,
we write auxiliary conditions as $\bm{m}_t$, which include language
embeddings, gripper-view observations, and other task-dependent metadata. Each
model uses the relevant subset of these conditions. We formulate the models as:
\begin{equation}
\begin{alignedat}{2}
\text{Policy:}\quad
\hat{\bm{a}}_{t:t+H-1}
&= \pi_\theta(\bm{s}_{t-H'+1:t}, \bm{z}_{t-H'+1:t}; \bm{m}_t^\pi), \\
\text{Dynamics:}\quad
\hat{\bm{o}}_{t+H}
&= f_\phi(\bm{o}_{t-H'+1:t}, \bm{a}_{t:t+H-1}; \bm{m}_t^f), \\
\text{Value:}\quad
\hat{V}_t
&= V_\psi(\bm{o}_t; \bm{m}_t^V).
\end{alignedat}
\label{eq:modeling_action_dynamics_value}
\end{equation}

$\pi_\theta$ predicts an action chunk from the history wrist state $\bm{s}_{t-H'+1:t}$ and the visual-token history $\bm{z}_{t-H'+1:t}$, conditioned on
policy-side auxiliary conditions $\bm{m}_t^\pi$ such as language embeddings~\cite{raffel2020exploring}.
The dynamics model $f_\phi$ and value model $V_\psi$ operate on the proposed world state $\bm{o}_t=[\bm{z}_t,\bm{P}_t]$.
Given world-state history and a candidate action chunk, $f_\phi$ predicts the future state $\hat{\bm{o}}_{t+H}$, while $V_\psi$ maps $\bm{o}_t$ to a scalar value. We omit auxiliary conditions $\bm{m}_t^\pi$, $\bm{m}_t^f$, and $\bm{m}_t^V$ in later equations when they are unchanged.

\paragraph{Model Pretraining.}
We first pretrain the policy, dynamics, and value models on human videos. The policy and dynamics models use the flow-matching paradigm introduced in Sec.~\ref{sec:preliminaries}, while the value model is optimized using a standard regression objective. Concretely, under the \emph{x-prediction} parameterization, the policy and dynamics networks directly regress clean targets from noisy samples at a randomly sampled timestamp $\tau \in [0,1]$.
Let ${\bm{a}}_{t:t+H-1}$, ${\bm{o}}_{t+H}$, and $V_t$ denote the corresponding supervision
targets. The losses are:
\begin{equation}
\label{eqn:humanpretrain}
\begin{aligned}
\mathcal{L}_{\pi}
&=
\left\|
\hat{\bm{a}}_{t:t+H-1}^{\,\tau}
-
\bm{a}_{t:t+H-1}
\right\|_2^2, \\
\mathcal{L}_{f}
&=
\left\|
\hat{\bm{o}}_{t+H}^{\,\tau}
-
\bm{o}_{t+H}
\right\|_2^2, \\
\mathcal{L}_{V}
&=
\left\|
\hat{V}_t
-
V_t
\right\|_2^2.
\end{aligned}
\end{equation}

We then adapt the dynamics and value models to robot experience and fine-tune the policy for each target task.

\begin{figure*}[t]
\centering
\includegraphics[width=0.85\linewidth]{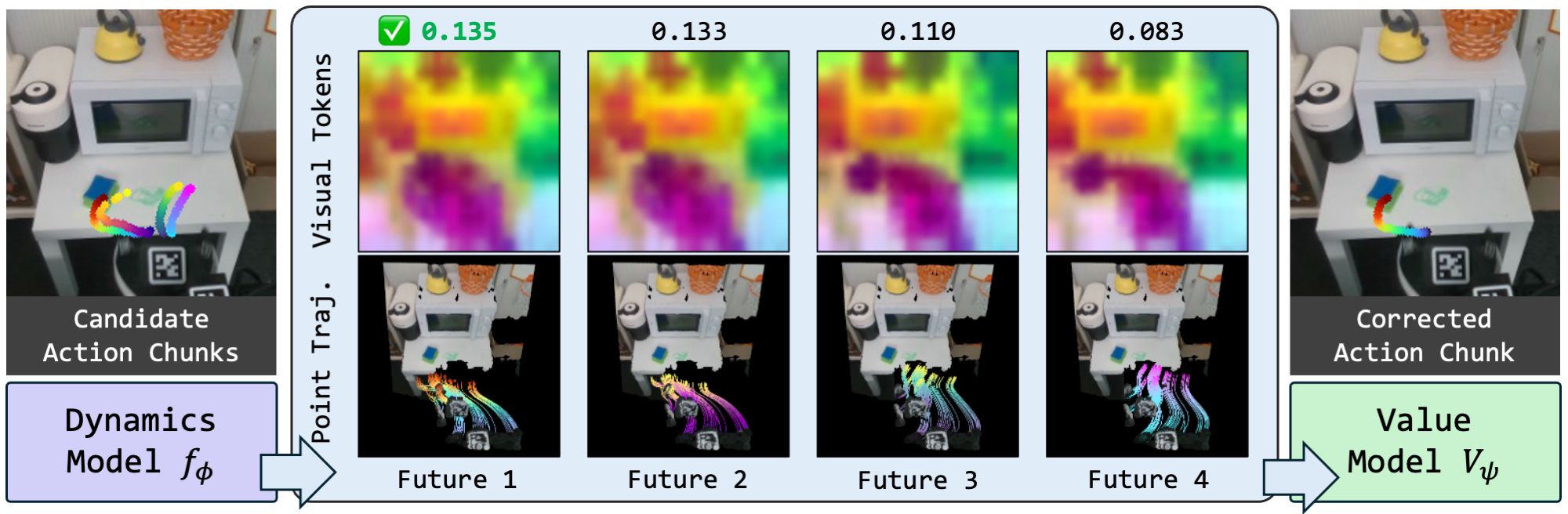}
\caption{
Given a failed state, DGAC samples candidate actions (left), rolls them out with the dynamics model, scores them with the value model, and selects the highest-value one (right) as corrective supervision.
}
\label{fig:DGAC}

\end{figure*}

\paragraph{Robot Adaptation of Dynamics and Value Models.}
The initial robot adaptation phase fine-tunes the pretrained dynamics and value models on task-agnostic robot rollouts. This adaptation addresses embodiment differences and exposes the models to robot-specific transitions and failures that are underrepresented in human videos.
To collect these rollouts without teleoperation, a VLM~\cite{openai2024gpt4o} proposes atomic interaction instructions (\textit{e.g.}, close the drawer). A pretrained affordance model~\cite{chen2025vidbot} proposes contact regions, and a motion planner moves the gripper to a selected region while avoiding collisions. The frozen human-pretrained policy then generates a post-contact trajectory, which the robot executes open-loop through its low-level controller. This separates contact acquisition from interaction generation, compensating for imprecise contact localization in the human-pretrained policy.
The resulting transitions and rollout outcomes supervise dynamics and value adaptation; the policy is not updated in this phase.
During subsequent task-level self-improvement, the dynamics and value models are further updated on rollouts from the current task policy (Sec.~\ref{sec:policy_improvement}).

\paragraph{Task-Specific Policy Fine-Tuning.}
For each target task, we fine-tune the human-pretrained policy on a small set of teleoperated demonstrations $\mathcal{D}_{\mathrm{tele}}$ using the policy loss $\mathcal{L}_{\pi}$. The resulting task policy serves as the starting point for iterative policy improvement.
The robot policy model is additionally conditioned on gripper-view observations through $\bm{m}_t^\pi$ and on an improvement indicator $I_t \in \{0,1\}$ used later for advantage-conditioned policy extraction:
\(
\hat{\bm{a}}_{t:t+H-1}
=
\pi_\theta\!\left(\bm{s}_{t-H'+1:t},\,\bm{z}_{t-H'+1:t};\,\bm{m}_t^\pi,\,I_t\right).
\)
During this task-specific fine-tuning, we set $I_t=1$ for all samples in
$\mathcal{D}_{\mathrm{tele}}$.

\vspace{-5pt}
\subsection{Dynamics-Guided Action Correction}
\label{sec:dgac}

Building on the task policy and the adapted dynamics and value models, we introduce \emph{Dynamics-Guided Action Correction} (DGAC) to construct corrective supervision from failed rollouts. DGAC uses successful experience to guide alternative action generation and evaluates the candidates with the learned models, as summarized in Alg.~\ref{alg:correction}.

\SetKwInput{KwRequire}{Require}
\newcommand{\GrayNl}[1]{\footnotesize\textcolor{gray}{#1}}
\begin{algorithm}[t]
\small
\SetNlSty{GrayNl}{}{}
\SetAlgoNoEnd
\SetAlgoNoLine
\DontPrintSemicolon
\caption{Dynamics-Guided Action Correction ({DGAC})}
\label{alg:correction}

\KwRequire{\begin{tabular}[t]{@{}l@{}}Policy $\pi_\theta$, dynamics model $f_\phi$, value \\model $V_\psi$, thresholds $\delta_{\rm v}, \delta_{\rm s}$\end{tabular}}
\KwIn{Failed state $\bm{o}_t$, successful dataset $\mathcal{D}_{\mathrm{succ}}$}
\KwOut{Corrected action $\bm{a}_{t:t+H-1}^\star$ or $\varnothing$}

Find $\mathcal{O}_t^{(k)}\subset\mathcal{D}_{\mathrm{succ}}$ and select $\tilde{\bm{o}}_t$ by Eq.~\ref{eqn:retrieval}\;
\If{$|V_\psi(\tilde{\bm{o}}_t)-V_\psi(\bm{o}_t)|>\delta_{\rm v}$ \KwSty{or} $\operatorname{sim}(\bm{o}_t,\tilde{\bm{o}}_t)<\delta_{\rm s}$}{
    \Return{$\varnothing$}
}

Sample action candidates
$\{\bm{a}_{t:t+H-1}^{(n)}\}_{n=1}^{N}$ via Eq.~\ref{eqn:comp_fm}\;

\For{$n = 1$ \KwTo $N$}{
    $\hat{\bm{o}}_{t+H}^{(n)} \leftarrow f_\phi(\bm{o}_{t-H'+1:t}, \bm{a}_{t:t+H-1}^{(n)})$\;
    $V_{t+H}^{(n)} \leftarrow V_\psi(\hat{\bm{o}}_{t+H}^{(n)})$\;
}

$n^\star \leftarrow \operatorname*{argmax}_n V_{t+H}^{(n)}$\;
$\bm{a}_{t:t+H-1}^\star \leftarrow \bm{a}_{t:t+H-1}^{(n^\star)}$\;

\Return{$\bm{a}_{t:t+H-1}^\star$}
\end{algorithm}

\begin{figure*}[t]
\centering
\includegraphics[width=0.9\linewidth]{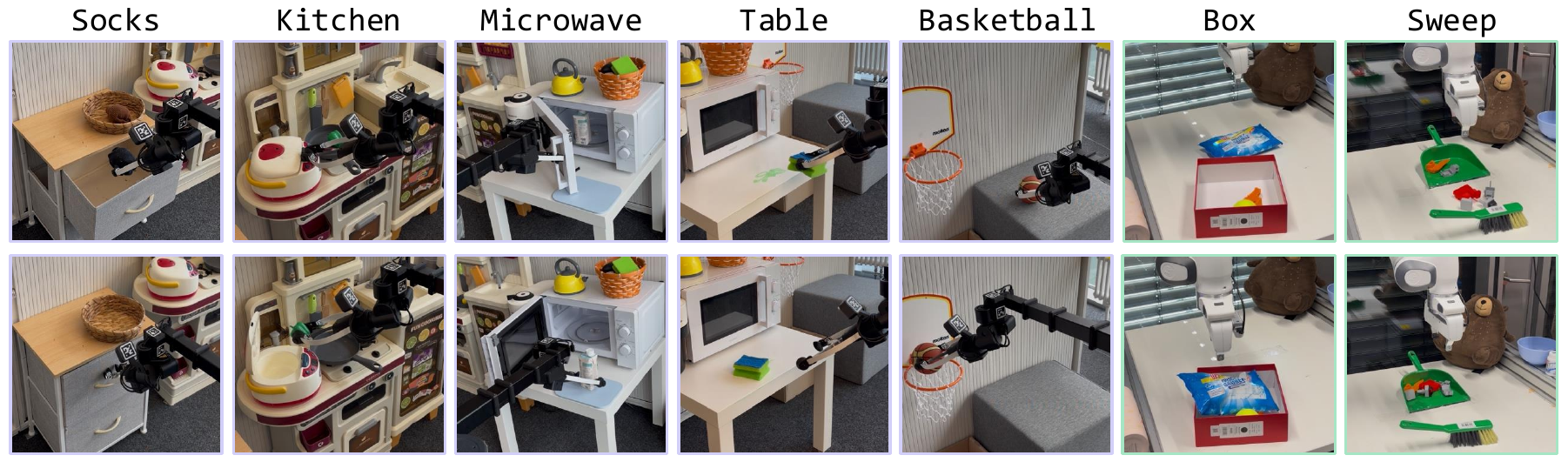}
\caption{
Seven real-world manipulation tasks across two robot platforms, spanning reaching, grasping, articulated-object manipulation, placement, and task-conditioned sequencing. Top row: intermediate states during task execution. Bottom row: final states upon task completion.
}
\label{fig:tasks}

\end{figure*}
The central idea is that many states from failed rollouts are \emph{recoverable} if they can be matched to successful interactions with similar predicted progress values and compatible scene configurations.
Rather than copying such references, DGAC uses them to \emph{steer} action generation and relies on the dynamics and value models to select the candidate predicted to best improve future task progress.
Let $\mathcal{D}_{\mathrm{succ}}$ and $\mathcal{D}_{\mathrm{fail}}$ denote successful and failed rollout data, respectively. Given a failed state $\bm{o}_t$ from $\mathcal{D}_{\mathrm{fail}}$, we retrieve the steering reference in two stages: we first retrieve successful states with similar predicted progress values, then select the most visually compatible one:
\begin{equation}
\label{eqn:retrieval}
\begin{aligned}
\mathcal{O}_t^{(k)} &= \operatorname{TopK}_{\bm{o}'\in\mathcal{D}_{\mathrm{succ}}}
\!\left(-|V_\psi(\bm{o}')-V_\psi(\bm{o}_t)|\right), \\
\tilde{\bm{o}}_t &= \operatorname{\argmax}_{\bm{o}'\in\mathcal{O}_t^{(k)}}
\operatorname{sim}(\bm{o}_t,\bm{o}').
\end{aligned}
\end{equation}
Here, $\operatorname{sim}$ denotes cosine similarity between the visual features of the two states.
We generate corrections only if the retrieved reference satisfies both $|V_\psi(\tilde{\bm{o}}_t)-V_\psi(\bm{o}_t)| \leq \delta_{\mathrm{v}}$ and $\operatorname{sim}(\bm{o}_t,\tilde{\bm{o}}_t) \geq \delta_{\mathrm{s}}$.
These heuristic thresholds restrict correction to states with progress-aligned, compatible successful references, but do not guarantee task recovery.

Because the policy is trained with conditional flow matching, it defines a velocity field over noisy action chunks under a given policy context, denoted by $\bm{v}_\theta(\bm{a}^{\tau}_{t:t+H-1}, \tau; \cdot)$. Let $\bm{h}_t = [\bm{s}_{t-H'+1:t}, \bm{z}_{t-H'+1:t}, \bm{m}_t^\pi, I_t]$ and $\tilde{\bm{h}}_t = [\tilde{\bm{s}}_{t-H'+1:t}, \tilde{\bm{z}}_{t-H'+1:t}, \tilde{\bm{m}}_t^\pi, \tilde{I}_t]$ denote the policy contexts associated with the current state $\bm{o}_t$ and the retrieved reference $\tilde{\bm{o}}_t$, respectively. Here, $I_t$ and $\tilde{I}_t$ are the advantage labels of the current sample and the retrieved reference sample, respectively, defined by the thresholding rule in Sec.~\ref{sec:preliminaries}.
We compose the corresponding velocity fields with weight $w_n \sim \mathcal{U}(0.1,1)$ as:
\begin{equation}
\label{eqn:comp_fm}
\begin{aligned}
\bm{v}^{(n)}_\theta
&=
\bm{v}_\theta(\bm{a}^{\tau}_{t:t+H-1}, \tau; \bm{h}_t) \\
&+
w_n \Big[
\bm{v}_\theta(\bm{a}^{\tau}_{t:t+H-1}, \tau; \tilde{\bm{h}}_t)
-
\bm{v}_\theta(\bm{a}^{\tau}_{t:t+H-1}, \tau; \bm{h}_t)
\Big].
\end{aligned}
\end{equation}
Integrating $\bm{v}^{(n)}_\theta$ from Gaussian noise yields a candidate action chunk $\bm{a}_{t:t+H-1}^{(n)}$. 
Because $w_n<1$, this composition retains a component conditioned on the current observation while steering generation toward the retrieved successful reference; sampling $w_n$ preserves diversity.
Repeating this process yields $N$ candidate action chunks $\{\bm{a}_{t:t+H-1}^{(n)}\}_{n=1}^N$.
As shown in Fig.~\ref{fig:DGAC}, we predict and score each candidate's outcome, then select the highest-value action chunk:
\begin{equation}
\label{eqn:dgac_evaluation}
\begin{aligned}
\hat{\bm{o}}_{t+H}^{(n)}
&= f_\phi(\bm{o}_{t-H'+1:t},\,\bm{a}_{t:t+H-1}^{(n)}), \ V_{t+H}^{(n)} = V_\psi(\hat{\bm{o}}_{t+H}^{(n)}), \\
n^\star 
&= \operatorname*{argmax}_n V_{t+H}^{(n)}, \quad \quad \bm{a}_{t:t+H-1}^{\star} = \bm{a}_{t:t+H-1}^{(n^\star)}.
\end{aligned}
\end{equation}
We initialize $\mathcal{D}_{\mathrm{repair}}$ from $\mathcal{D}_{\mathrm{fail}}$ and replace the original actions with selected corrections that satisfy the reference checks.

\subsection{Iterative Policy Improvement}
\label{sec:policy_improvement}

\paragraph{Iterative Self-Improvement Loop.}
With the adapted models and the initialized task policy, we iteratively improve the robot policy through deployment and relabeling, as summarized in Fig.~\ref{fig:pipeline}.
At each iteration, we execute the current policy $\pi_\theta$ in the real environment to collect trajectories. We partition the collected trajectories into successful and failed datasets, denoted by $\mathcal{D}_{\mathrm{succ}}$ and $\mathcal{D}_{\mathrm{fail}}$, respectively. The successful dataset is initialized with $\mathcal{D}_{\mathrm{tele}}$.
We fine-tune the dynamics model $f_\phi$ and value model $V_\psi$ on all collected data so that they adapt to the policy-induced state distribution.
For failed states in $\mathcal{D}_{\mathrm{fail}}$, we invoke the DGAC module described in Sec.~\ref{sec:dgac} to turn failure-inducing actions into corrective training targets using $f_\phi$ and $V_\psi$, forming a repaired dataset $\mathcal{D}_{\mathrm{repair}}$.
Finally, we estimate advantages on $\mathcal{D}_{\mathrm{succ}} \cup \mathcal{D}_{\mathrm{repair}}$ and update $\pi_\theta$ through advantage-conditioned policy extraction for the next iteration.

\paragraph{Advantage-Conditioned Policy Update.}
We estimate the advantage
$\hat{A}_t$ on $\mathcal{D}_{\mathrm{succ}} \cup \mathcal{D}_{\mathrm{repair}}$ via Eq.~\ref{eqn:advantage}, using dynamics-predicted successor states for corrected samples, and update the policy with the advantage-conditioned policy
extraction objective of CFGRL~\cite{frans2025diffusion}, described in Sec.~\ref{sec:preliminaries}.

\paragraph{Optimization Objective.}
For training on robot data, we use the same objectives as in Eq.~\ref{eqn:humanpretrain} for all three models. We further augment $\mathcal{L}_{V}$ with a temporal-difference (TD) term~\cite{sutton1988learning} following RISE~\cite{yang2026rise}: 
\begin{equation}
\label{eqn:tdloss}
\begin{aligned}
\mathcal{L}_{\mathrm{TD}}=\|\hat{V}_t-(r_t+\gamma\hat{V}_{t+1})\|_2^2,
\end{aligned}
\end{equation}
where $\gamma$ is the discount factor.
This objective refines the value estimate to better distinguish between successes and failures.

\subsection{Implementation Details}
\label{sec:impl_details}
We pretrain our models on approximately one million samples from HOI4D~\cite{liu2022hoi4d}, Arti4D~\cite{werby2025articulated}, and EgoDex~\cite{hoque2025egodex}. DINO-v3 extracts visual features, and TAPIP3D produces 3D point tracks using dataset camera parameters and depth; for EgoDex, depth is estimated with DepthAnything-v3~\cite{lin2025depth}. We obtain wrist poses from dataset annotations where available; otherwise, HaMeR~\cite{pavlakos2024reconstructing} and point tracking provide translations and temporally smoothed orientations. We retain a two-hand, 20D action format from human pretraining: each hand uses 3D translation, a continuous 6D rotation representation~\cite{zhou2019continuity}, and a scalar closure value. For the single-arm robots evaluated here, we duplicate the robot action into the unused hand slot for compatibility; this duplicate is ignored by the controller. Initial robot adaptation uses approximately 400 autonomous episodes collected in three hours with the procedure in Sec.~\ref{sec:method_models}; 72.8\% of these episodes succeed.

The policy and dynamics models each use 16 transformer layers, 12 attention heads, and width 384; the value model uses 8 layers, 8 heads, and the same width. Action chunks contain $H=30$ steps. Pretraining, initial dynamics/value adaptation, and per-round dynamics/value fine-tuning use 250k, 80k, and 8k updates with batch sizes 20, 20, and 12, respectively, using AdamW~\cite{loshchilov2017decoupled} with a learning rate of $10^{-4}$. Task-policy initialization and each policy update use 30k training steps with batch size 32. We weight the visual dynamics loss by 100 and the TD loss by 10, and set the advantage-label threshold to the empirical 70th percentile.

\begin{figure*}[t]
\centering
\includegraphics[width=0.95\linewidth]{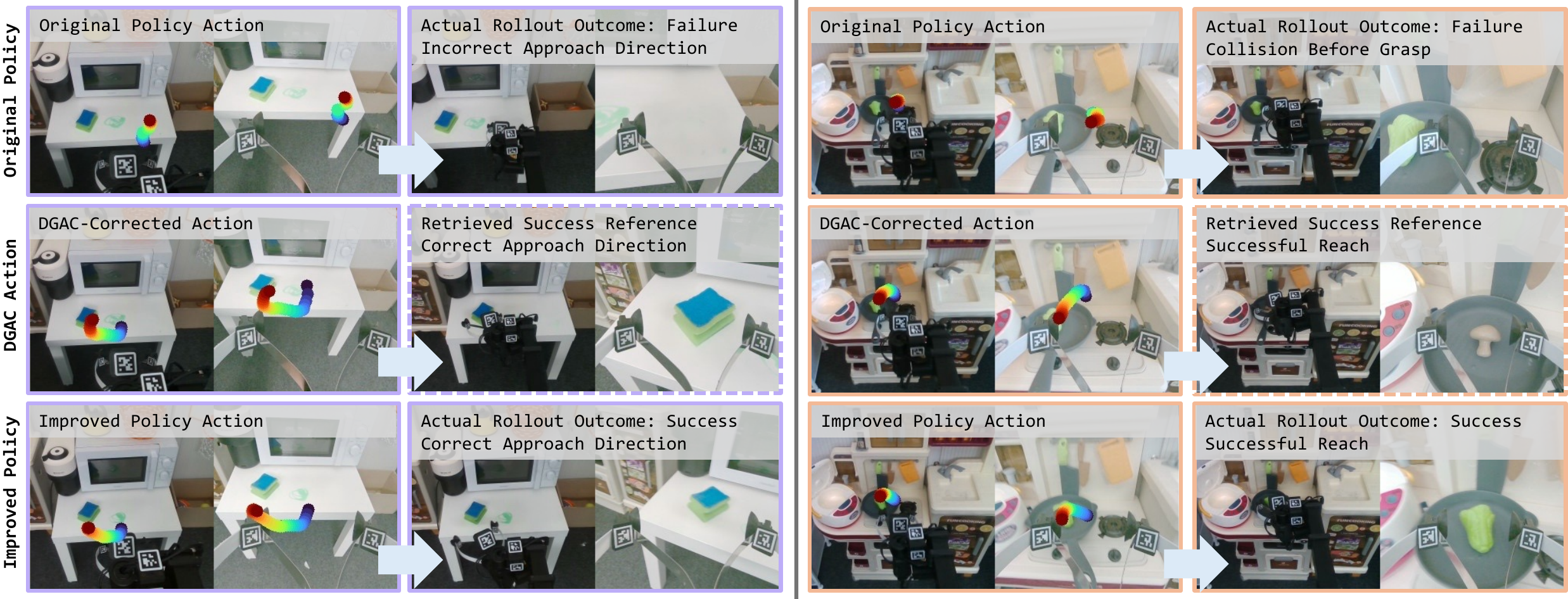}
\caption{DGAC constructs corrective supervision from failed rollouts. Original failure-inducing actions and their outcomes are shown at the top, corrected actions and retrieved successful references in the middle, and real rollouts after training on corrections at the bottom.}
\label{fig:qualitative}
\vspace{-2pt}
\end{figure*}

\section{Experiments}
\label{sec:experiments}
\setlength{\textfloatsep}{10pt}
\setlength{\floatsep}{10pt}

We evaluate robot self-improvement through model-generated corrective supervision, examining both the human-video priors and their use in DGAC. Our
  experiments address four questions:
   (1) Does the complete framework improve real-world manipulation, and does the improvement hold across rounds, embodiments, and policy backbones?
  (Sec.~\ref{sec:exp_overall})
   (2) Are DGAC corrections effective, both when executed directly from failed states and when used as policy-training targets?
  (Sec.~\ref{sec:exp_correction})
   (3) Which failures does corrective supervision cover, and does it bias the advantage estimates used for training? (Sec.~\ref{sec:exp_correction})
   (4) How do human-video priors and world-state representations contribute to improvement? (Sec.~\ref{sec:exp_priors})

\subsection{Experimental Setup}
\paragraph{Evaluation Setup.}
We conduct baseline comparisons and ablation studies using Hello Robot Stretch 3.
Our main benchmark includes five manipulation tasks, with each policy evaluated over 15 independent trials per task.

To assess cross-embodiment applicability, we further test our method on two additional tasks using a Franka Panda robot.
All tasks are visualized in Fig.~\ref{fig:tasks}.
Training proceeds in four stages. We pretrain all three models on approximately one million human-video samples, then adapt the dynamics and value models on autonomous robot episodes (Sec.~\ref{sec:impl_details}). For each task, we fine-tune the policy on 25 teleoperated demonstrations, and then run two self-improvement rounds that update all three models. All policy improvement methods collect 20 episodes per round. During improvement, humans only reset scenes and label rollout outcomes; they provide no corrective actions or interventions. \textbf{Ours} denotes the same final policy in all tables.

\paragraph{Baselines.}
Expert BC is the human-pretrained policy fine-tuned only on the 25 demonstrations and serves as the initialization for policy improvement methods. AWR~\cite{peng2019AWR} weights samples by clipped exponential advantages. RECAP~\cite{intelligence2025pi} conditions the policy on binary advantage labels with a 70th-percentile cutoff. For RISE~\cite{yang2026rise}, we fine-tune the provided video-model checkpoint on our robot rollouts to synthesize policy-training data. RECAP and RISE receive no human corrective interventions.
We reimplement SWIM~\cite{mendonca2023SWIM} with our architectural configurations and its structured action space, using our goal retrieval and ranking the predicted outcomes by feature distance to the reference. We adapt LPB~\cite{sun2025lpb} to flow matching by guiding the velocity field with gradients of latent distance to the goal; our value model accelerates goal retrieval.

\begin{table}[t!]
\centering
\setlength{\tabcolsep}{4pt}
\fontsize{7}{8}\selectfont
\renewcommand{\arraystretch}{0.95}
\begin{tabularx}{\resulttablewidth}{@{}r|*{5}{Y}|Y@{}}
\toprule
\textbf{Method}
    & \textbf{\texttt{T1}}
    & \textbf{\texttt{T2}}
    & \textbf{\texttt{T3}}
    & \textbf{\texttt{T4}}
    & \textbf{\texttt{T5}}
    & \textbf{Avg.} \\
\midrule
Expert BC
    & 33.3
    & 46.7
    & 46.7
    & 40.0
    & 40.0
    & 41.3 \\
\midrule
SWIM~\cite{mendonca2023SWIM}
    & 53.3
    & 40.0
    & 46.7
    & 46.7
    & 60.0
    & 49.3 \\
LPB~\cite{sun2025lpb}
    & 53.3
    & 40.0
    & 66.7
    & \second{80.0}
    & 60.0
    & 60.0 \\
AWR~\cite{peng2019AWR}
    & 53.3
    & {53.3}
    & \best{80.0}
    & 60.0
    & 53.3
    & 60.0 \\
RECAP$^\dagger$~\cite{intelligence2025pi}
    & \second{60.0}
    & {53.3}
    & \second{73.3}
    & 53.3
    & {66.7}
    & {61.3} \\
RISE$^\dagger$~\cite{yang2026rise}
    & \best{86.7}
    & \second{60.0}
    & \second{73.3}
    & \second{80.0}
    & \second{80.0}
    & \second{76.0} \\

\midrule
\textbf{Ours}
    & \best{86.7}
    & \best{80.0}
    & \best{80.0}
    & \best{93.3}
    & \best{86.7}
    & \best{85.3} \\
\bottomrule
\end{tabularx}
\caption{Benchmark evaluated on success rate (\%).
    \textbf{\texttt{T1}}: {Socks},
    \textbf{\texttt{T2}}: {Kitchen},
    \textbf{\texttt{T3}}: {Microwave},
    \textbf{\texttt{T4}}: {Table},
    \textbf{\texttt{T5}}: {Basketball}.
    $^\dagger$: no human-intervened corrections. Dark/light green: best/second best.
}
\label{tab:benchmark}

\end{table}

\subsection{Overall Policy Improvement}
\label{sec:exp_overall}
\paragraph{Does Our Framework Improve Real-World Manipulation?}
Tab.~\ref{tab:benchmark} shows that our framework achieves 85.3\% average success across five Stretch tasks, versus 41.3\% for Expert BC. AWR and RECAP achieve 60.0\% and 61.3\%. They weight or label collected data by advantage but cannot provide better actions at failed states, so failed rollouts mainly teach the policy what to avoid. SWIM and LPB achieve 49.3\% and 60.0\%. Both select or steer actions using latent distance to a retrieved goal, which is likely less informative than a learned value in these multi-stage tasks. RISE is the strongest baseline at 76.0\%. It adds rollouts synthesized by a video world model, while DGAC constructs corrections at recorded failure states. Sec.~\ref{sec:exp_correction} further compares supervision-construction procedures using our dynamics and value models. Fig.~\ref{fig:qualitative} illustrates corrections and resulting policy rollouts.

\begin{figure*}[t]
    \centering
    \includegraphics[width=0.95\textwidth]{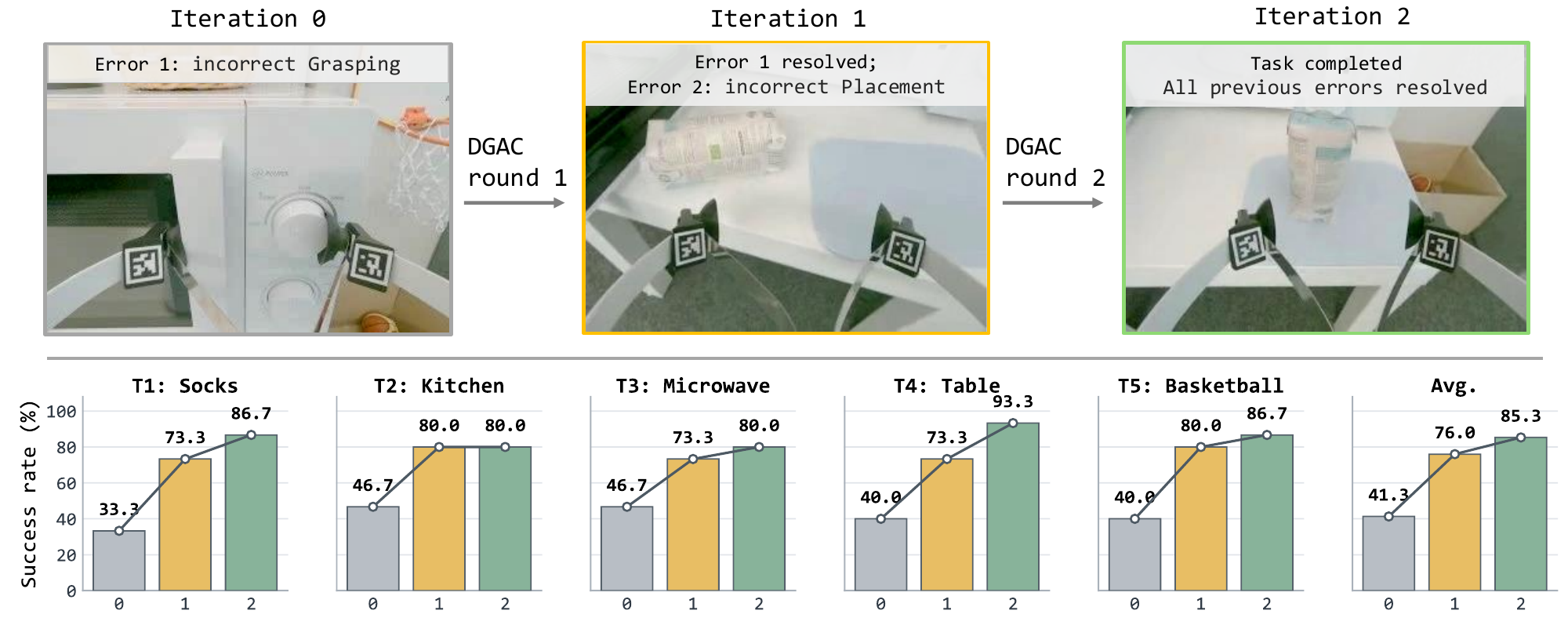}
    \caption{\textbf{Policy performance across DGAC rounds.} With 20 newly collected episodes per round, average success increases from 41.3\% to 76.0\% after one round and 85.3\% after two. Top: a \texttt{microwave} example shows successive rounds correcting failures, enabling task completion. Bottom: success rates (\%) on five Stretch tasks and their average; iteration 0 denotes the initial Expert BC policy.}
    \label{fig:per_iteration}
\end{figure*}

\paragraph{How Does Policy Performance Evolve across DGAC Rounds?}
DGAC rapidly improves the initialized policy with only 20 newly collected rollout episodes per round (Fig.~\ref{fig:per_iteration}). Across five Stretch tasks, average success increases from 41.3\% to 76.0\% after a single round, a gain of 34.7 percentage points. A second round raises the average to 85.3\%, without reducing success on any task. The \texttt{microwave} example illustrates how successive rounds address failures exposed by the evolving policy: the first round resolves an incorrect grasp, allowing execution to reach a later placement failure, which the second round resolves. This progression highlights the role of repeated correction: as the policy advances further through a task, DGAC provides corrective supervision for the failures it encounters. These results suggest that, over the two evaluated rounds, DGAC addresses newly encountered failures while improving overall performance.
\begin{table}[t]
\centering

\setlength{\tabcolsep}{4pt}
\fontsize{7}{8}\selectfont
\renewcommand{\arraystretch}{0.95}
\begin{tabularx}{\resulttablewidth}{@{}r|*{5}{Y}|Y@{}}
\toprule
\textbf{Method}
& \textbf{\texttt{T1}}
& \textbf{\texttt{T2}}
& \textbf{\texttt{T3}}
& \textbf{\texttt{T4}}
& \textbf{\texttt{T5}}
& \textbf{Avg.} \\
\midrule
$\pi_{0.5}$ + SFT
& 60.0
& 66.7
& \second{80.0}
& 40.0
& 66.7
& 62.7 \\
$\pi_{0.5}$ + RECAP$^\dagger$
& 66.7
& 66.7
& 73.3
& 53.3
& \second{80.0}
& 68.0 \\
$\pi_{0.5}$ + DGAC
& \best{93.3}
& \best{86.7}
& \best{86.7}
& \second{86.7}
& \best{86.7}
& \best{88.0} \\
\midrule
\textbf{Ours}
& \second{86.7}
& \second{80.0}
& \second{80.0}
& \best{93.3}
& \best{86.7}
& \second{85.3} \\
\bottomrule
\end{tabularx}

\caption{Success rates (\%) of the baseline policy $\pi_{0.5}$ under different policy improvement schemes. $^\dagger$: no human-intervened corrections.}
\label{tab:ablation_pi0_dgac}

\end{table}

\paragraph{Does the Framework Apply to Another Embodiment?}
Following the Stretch protocol, we further test our self-improvement pipeline on the Franka Panda. Our method achieves 73.3\% on \texttt{Box} and 66.7\% on \texttt{Sweep}, compared with 46.7\% and 26.7\% for Expert BC. The two-task average increases from 36.7\% to 70.0\%. Across all seven tasks, the average is 81.0\% for our method and 40.0\% for Expert BC. These results show that the shared modeling interfaces and correction procedure support improvements on both platforms.

\paragraph{Can DGAC Improve a Different Policy Backbone?}
We apply DGAC to $\pi_{0.5}$~\cite{physicalintelligence2025pi05}, a policy pretrained on large-scale robot data, while keeping the human-pretrained dynamics and value models of our framework. We fine-tune $\pi_{0.5}$ with LoRA in the same action space as our policy, so DGAC generates and evaluates its corrections as in Sec.~\ref{sec:dgac}. Since $\pi_{0.5}$ is not conditioned on advantage labels, RECAP and DGAC update it only on samples with $I_t=1$. On the five Stretch tasks, supervised fine-tuning (SFT) on the 25 demonstrations yields 62.7\% success. Training additionally on rollout data with RECAP improves the average to 68.0\%, and adding DGAC corrections raises it to 88.0\%, 20.0 percentage points above RECAP (Tab.~\ref{tab:ablation_pi0_dgac}). The correction procedure therefore also works with a second policy backbone: a robot-pretrained policy gains from human-pretrained dynamics and value models, consistent with the DROID comparison in Sec.~\ref{sec:exp_priors}.

\subsection{Effectiveness and Coverage of Corrective Supervision}
\label{sec:exp_correction}
We evaluate DGAC corrections in two ways: executing them from failed states, which tests the corrections themselves, and distilling them into the policy, which tests their value as training targets. We then analyze the predicted outcomes, which failures contribute to corrective supervision, the limits of its coverage, and the estimated advantages.

\paragraph{Does DGAC Enable Recovery from Failed States?}\mbox{}

\begin{wraptable}{R}{0.51\columnwidth}
\vspace{-0.6em}
\centering
\setlength{\tabcolsep}{5pt}
\fontsize{7}{8}\selectfont
\renewcommand{\arraystretch}{0.95}
\begin{tabular}{@{}r|ccc|c@{}}
\toprule
\textbf{Rank}
& \textbf{\texttt{T1}}
& \textbf{\texttt{T4}}
& \textbf{\texttt{T5}}
& \textbf{Avg.} \\
\midrule
{Top}
& \best{70.0}
& \best{40.0}
& \best{80.0}
& \best{63.3} \\
Med.
& \second{60.0}
& \second{20.0}
& \second{20.0}
& \second{33.3} \\
Low
& 10.0
& 0.0
& 0.0
& 3.3 \\
\bottomrule
\end{tabular}
\caption{Success rates (\%) from ten distinct failed states per task.}
\label{tab:dgac_recovery}
\vspace{-1.2em}
\end{wraptable}

\noindent We first execute DGAC corrections directly, without any policy update. We randomly select ten distinct failed states for each of three tasks (\texttt{T1}, \texttt{T4}, \texttt{T5}). From each state, we separately execute the top-, median-, and lowest-ranked candidate once, resetting the scene and robot to the failed state before each execution. Each trial executes a single correction chunk and then lets the original policy continue until the episode ends, so all three ranks are evaluated at the same 30 failed states. We do not report re-executing the original policy from these states, since in our experiments it typically reproduced the failure.
Recovery rates (Tab.~\ref{tab:dgac_recovery}) decrease from 63.3\% for the top-ranked candidate to 33.3\% for the median-ranked and 3.3\% for the lowest-ranked one (19, 10, and 1 successes out of 30 trials per rank). This shows that candidates ranked higher by the dynamics and value models are more likely to recover when executed.
\begin{table}[t]
\centering
\setlength{\tabcolsep}{4pt}
\fontsize{7}{8}\selectfont
\renewcommand{\arraystretch}{0.95}

\begin{tabularx}{\resulttablewidth}{@{}r|*{5}{Y}|Y@{}}
\toprule
\textbf{Variant}
& \textbf{\texttt{T1}}
& \textbf{\texttt{T2}}
& \textbf{\texttt{T3}}
& \textbf{\texttt{T4}}
& \textbf{\texttt{T5}}
& \textbf{Avg.} \\
\midrule
w/o {DGAC} [V1]      
& 66.7 
& 60.0 
& 60.0 
& 66.7
& 60.0 
& 62.7 \\
Reference [V2] 
& 40.0 
& 53.3 
& \second{73.3} 
& 60.0 
& 66.7 
& 58.7 \\
RISE-style$^\ddagger$ [V3]
& \second{73.3}
& \second{66.7}
& 60.0
& \second{73.3}
& \second{80.0}
& \second{70.7} \\
\midrule
Rand. Val. [V4] 
& 33.3 
& 53.3 
& 60.0 
& 46.7 
& 66.7 
& 52.0 \\
Med. Val. [V5]       
& 46.7 
& 60.0 
& 60.0 
& {66.7} 
& 53.3 
& 57.3 \\
Min. Val. [V6]        
& 26.7 
& 53.3 
& 53.3 
& 53.3 
& 53.3 
& 48.0 \\
VLM Val. [V7]
& 60.0
& 53.3
& 66.7
& \second{73.3}
& 66.7
& 64.0 \\
\midrule
\textbf{Ours}   
& \best{86.7}
& \best{80.0}
& \best{80.0}
& \best{93.3}
& \best{86.7}
& \best{85.3} \\

\bottomrule
\end{tabularx}
\caption{Effect of corrective supervision construction on success rate (\%).
All variants share the two rounds of robot rollouts and the models of Ours, and differ only in how corrective actions are constructed from failed states. Top: target generation; middle: candidate selection from the DGAC pool.
$^\ddagger$: RISE-style pipeline that synthesizes rollouts with our dynamics and value models, instead of with RISE models.}
\vspace{-5pt}
\label{tab:ablation_dgac}

\end{table}%
\begin{figure*}[t]
\centering
\includegraphics[width=0.95\linewidth]{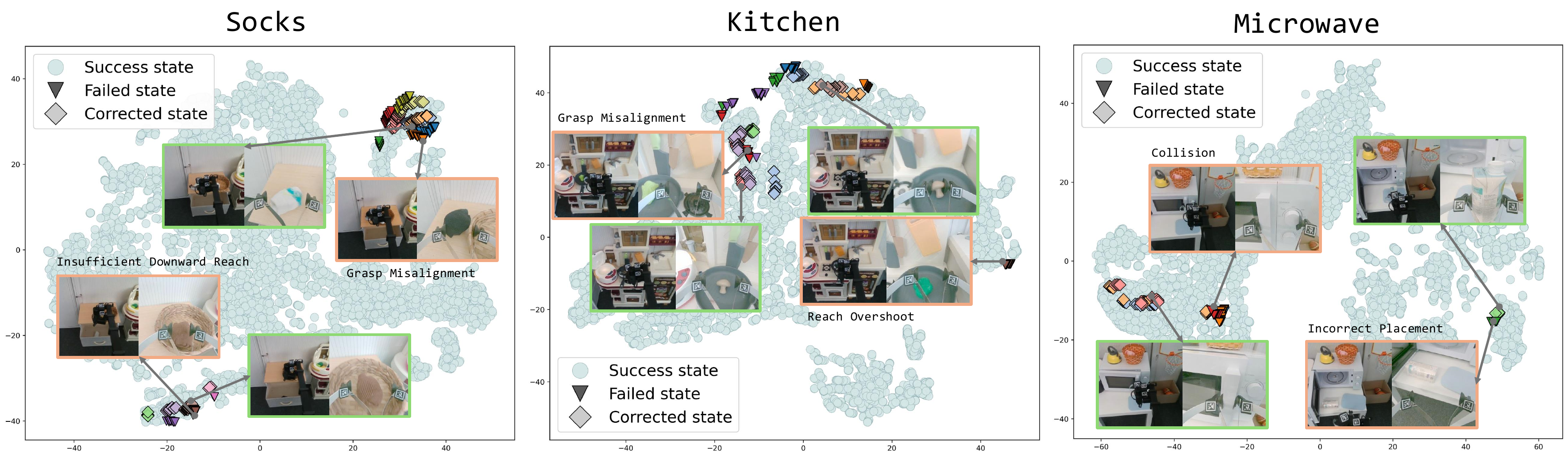}
\caption{
t-SNE visualization of state dynamics embeddings across three real-world tasks. Light blue circles denote successful states, inverted triangles denote states from failed rollouts, and diamonds denote predicted future states after
DGAC correction. For representative predictions, we show their nearest successful neighbors. These examples illustrate candidate corrections for grasp misalignment, insufficient reach, and incorrect placement whose predicted outcomes lie near successful states in the embedding.
}
\label{fig:tsne_dgac}
\end{figure*}

\paragraph{Do DGAC Corrections Yield a Better Policy?}
We next distill corrections into the policy. Tab.~\ref{tab:ablation_dgac} compares ways of constructing training targets from the same robot rollouts: the two rounds collected in our main experiment (Fig.~\ref{fig:per_iteration}), the first by Expert BC and the second by the first-round DGAC policy. All variants share these rollouts, the policy backbone, and the dynamics and value models. They differ only in the training targets constructed for failed states: how candidate actions are generated (V1--V3) and how a correction is selected from the DGAC candidate pool (V4--V7). Because the rollouts are fixed in advance, this comparison isolates the effect of the constructed supervision.

\emph{Target generation.} Training without corrections (V1), i.e., RECAP$^\dagger$ on the shared rollouts, yields 62.7\%, consistent with Tab.~\ref{tab:benchmark} (61.3\%). Directly relabeling failed states with the retrieved reference action (V2), without adapting it to the current scene, yields 58.7\%. A RISE-style pipeline~\cite{yang2026rise} (V3), which synthesizes rollouts of the current policy with our dynamics model and keeps those with high predicted value, reaches 70.7\%. DGAC reaches 85.3\% by using the reference to guide proposals conditioned on both contexts and evaluating their predicted consequences from the failed state with the learned dynamics and value models. 

\begin{figure}[t]
\centering
\includegraphics[width=\linewidth]{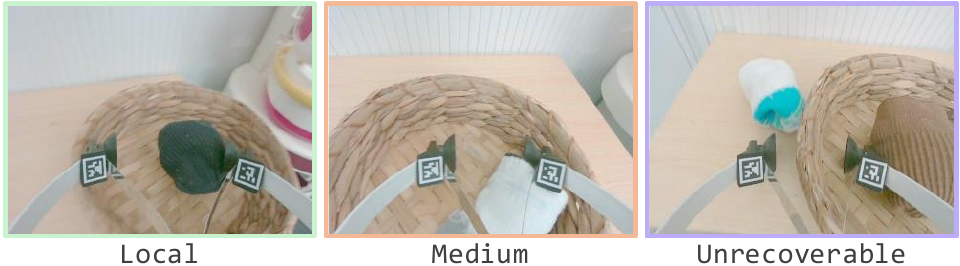}
\caption{Failure categories at error onset. \emph{Local}~/~\emph{Medium}: recoverable small/large deviations from successful execution. \emph{Unrecoverable}: irreversible scene changes that prevent recovery.}
\label{fig:failure_taxonomy}
\end{figure}
\begin{table}[!t]
\centering
\setlength{\tabcolsep}{4pt}
\fontsize{7}{8}\selectfont
\renewcommand{\arraystretch}{1.05}
\begin{tabular}{@{}l|r|cc|c@{}}
\toprule
\textbf{Category}
& \textbf{Episodes}
& \makecell{\textbf{Accepted}\\\textbf{frames}}
& \makecell{\textbf{Frames used}\\\textbf{for training}}
& \makecell{\textbf{Contributing}\\\textbf{episodes}} \\
\midrule
Local & 14 & 68\% & 55\% & 14/14 \\
Medium & 15 & 38\% & 24\% & 10/15 \\
Unrecoverable & 8 & 23\% & 18\% & 3/8 \\
\bottomrule
\end{tabular}
\caption{Coverage of corrective supervision over 37 annotated failed episodes. Frame shares are computed from the error onset onward and pooled within each category. \emph{Accepted}: the generated correction satisfies both thresholds $\delta_{\mathrm{v}}$, $\delta_{\mathrm{s}}$. \emph{Used for training}: an accepted correction also passes the advantage threshold.}
\label{tab:failure_taxonomy}
\end{table}

\emph{Candidate selection.} With the DGAC candidate pool held fixed, random, median-value, and minimum-value selection (V4--V6) yield 52.0\%, 57.3\%, and 48.0\%. All three fall below training without corrections (V1, 62.7\%), indicating that candidate generation alone does not ensure useful supervision. The top/median/minimum ordering agrees with the direct-execution results in Tab.~\ref{tab:dgac_recovery}: higher-ranked candidates both lead to recovery more often and make better training targets. A VLM selector~\cite{openai2025gpt5} following SOAR~\cite{zhou2024autonomous} (V7) achieves a success rate of 64.0\%, lower than that achieved by dynamics/value-based selection from the same candidate pool. Fig.~\ref{fig:qualitative} shows corrected actions, their retrieved references, and rollouts of the policy after training on them.

\begin{figure*}[t]
\centering
\includegraphics[width=\linewidth]{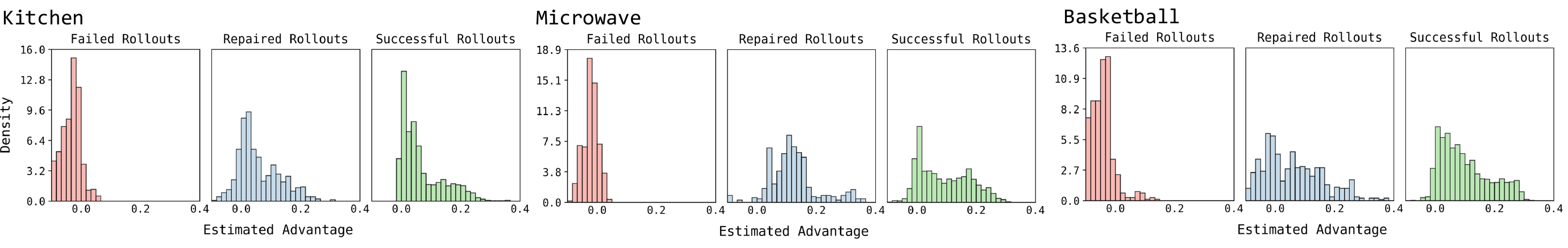}
\caption{
Estimated advantage distributions of failed, repaired, and successful samples on three tasks.
}
\vspace{-10pt}
\label{fig:adv_dist}
\end{figure*}

\paragraph{Do Predicted Corrections Move Toward Successful States?}
Fig.~\ref{fig:tsne_dgac} shows successful, failed, and DGAC-predicted states in the learned embedding for three tasks, with nearest successful neighbors for selected predictions. Predicted outcomes tend to lie closer to successful states. The examples include insufficient downward reach and grasp misalignment in \texttt{Socks}, grasp misalignment and reach overshoot in \texttt{Kitchen}, and collision and incorrect placement in \texttt{Microwave}.

\begin{table}[t]
\centering
\setlength{\tabcolsep}{4pt}
\fontsize{7}{8}\selectfont
\renewcommand{\arraystretch}{0.95}
\begin{tabularx}{\resulttablewidth}{@{}cc|*{5}{Y}|Y@{}}
\toprule
\multicolumn{2}{c|}{\textbf{Pretraining source}}
& \textbf{\texttt{T1}} & \textbf{\texttt{T2}} & \textbf{\texttt{T3}}
& \textbf{\texttt{T4}} & \textbf{\texttt{T5}} & \textbf{Avg.} \\
$\pi_\theta$ & $f_\phi,V_\psi$ & & & & & & \\
\midrule
DROID & DROID & 53.3 & 40.0 & \second{66.7} & 73.3 & \second{73.3} & 61.3 \\
DROID & Human & \second{66.7} & \second{60.0} & \second{66.7} & \second{80.0} & \second{73.3} & \second{69.3} \\
\midrule
Human & Human & \best{86.7} & \best{80.0} & \best{80.0} & \best{93.3} & \best{86.7} & \best{85.3} \\
\bottomrule
\end{tabularx}
\caption{Effect of different combinations of pretraining sources on final success rate (\%).}
\label{tab:pretraining_sources}

\end{table}
\paragraph{Which Failures Yield Corrective Supervision?}
We label 37 failed episodes by their state at error onset (Fig.~\ref{fig:failure_taxonomy}); two annotators label each episode and resolve disagreements by discussion. In \emph{local} and \emph{medium} failures, the task remains completable, and the end-effector deviation from successful execution is below or above a task-specific threshold, respectively (3~cm on average). In \emph{unrecoverable} failures, the scene has changed so that the task can no longer be completed. This gives 14 local, 15 medium, and 8 unrecoverable episodes. We apply DGAC to every frame from the error onset onward. A frame is \emph{accepted} if its generated correction satisfies both thresholds $\delta_{\mathrm{v}}$ and $\delta_{\mathrm{s}}$ (Sec.~\ref{sec:dgac}), and \emph{used for training} if its correction passes the advantage threshold.

Tab.~\ref{tab:failure_taxonomy} shows that DGAC primarily generates training corrections for failures close to successful execution. All 14 local episodes contribute training frames, and 55\% of their frames are used. For medium failures, 10 of 15 episodes contribute and 24\% of frames are used, since fewer frames find a compatible successful reference. This is the main limitation of the current correction procedure: correcting larger deviations requires broader coverage of successful experience. In unrecoverable episodes, the reference checks reject 77\% of frames, but 18\% still pass all checks and are used for training, as the value and similarity checks do not always detect that the scene has changed irreversibly. These frames make up only 6.95\% of all training corrections.

\paragraph{Do Corrections Inflate Estimated Advantages?}
Fig.~\ref{fig:adv_dist} shows the estimated advantages of failed, repaired, and successful samples on three tasks. The estimated advantages of failed samples concentrate near or below zero. After correction, the repaired samples shift into the range of successful samples without systematically exceeding it; only a small tail extends beyond that range in \emph{Microwave} and \emph{Basketball}. \begin{table}[t]
\centering
\setlength{\tabcolsep}{4pt}
\fontsize{7}{8}\selectfont
\renewcommand{\arraystretch}{0.95}
\begin{tabularx}{\resulttablewidth}{@{}r|*{5}{Y}|Y@{}}
\toprule
\textbf{Variant}
& \textbf{\texttt{T1}}
& \textbf{\texttt{T2}}
& \textbf{\texttt{T3}}
& \textbf{\texttt{T4}}
& \textbf{\texttt{T5}}
& \textbf{Avg.} \\
\midrule
Human 0\%
& 53.3
& \second{66.7}
& 53.3
& 40.0
& 46.7
& 52.0 \\
Human 50\%
& 60.0
& 53.3
& 60.0
& 60.0
& 66.7
& 60.0 \\
\midrule
Adapt. 0\%
& 60.0
& 53.3
& 60.0
& 73.3
& \second{80.0}
& 65.3 \\
Adapt. 25\%
& 60.0
& 60.0
& 66.7
& 73.3
& \second{80.0}
& 68.0 \\
Adapt. 50\%
& \second{66.7}
& \best{80.0}
& \second{73.3}
& \second{86.7}
& \best{86.7}
& \second{78.7} \\
\midrule
\textbf{Ours}
& \best{86.7}
& \best{80.0}
& \best{80.0}
& \best{93.3}
& \best{86.7}
& \best{85.3} \\
\bottomrule
\end{tabularx}
\caption{Effect of human-video pretraining and robot adaptation data scale on success rate (\%).}
\label{tab:ablation_human_pretrain}
\vspace{-10pt}
\end{table}
This suggests that corrections do not systematically shift the distribution from which the advantage labels are derived. Because these estimates come from the same models that select the corrections, they cannot by themselves show that corrections succeed; we verify corrections by direct execution in Tab.~\ref{tab:dgac_recovery}.

\subsection{Role of Human-Video
  Priors and World Representations}
\label{sec:exp_priors}
\paragraph{Does the Pretraining Source Matter?}
We replace human-video pretraining with DROID~\cite{khazatsky2024droid,huang2026pointworld}, a large-scale robot teleoperation dataset, while keeping the model architectures, downstream data protocol, DGAC procedure, training schedule, and pretraining dataset scale fixed. Pretraining all three models on DROID yields 61.3\% average success, compared with 85.3\% for human-video pretraining (Tab.~\ref{tab:pretraining_sources}). To isolate the correction side of the pipeline, we keep the DROID-pretrained policy and replace only the dynamics and value models with their human-pretrained counterparts. This raises success from 61.3\% to 69.3\%, improving or matching every task. Additionally using the human-pretrained policy, which both initializes the task policy and proposes DGAC candidates, reaches 85.3\%. These results suggest that human-video pretraining benefits both policy initialization and the dynamics and value models that construct corrections.

With the two corpora matched in scale, the performance gap suggests that the content of the pretraining data matters. Our human videos include many everyday objects, articulated objects, and household scenes, while teleoperated robot data are limited by the cost of collecting data on physical robots. This broader coverage is one possible explanation; the corpora also differ in viewpoint and task actions, which this comparison does not disentangle.

\paragraph{How Do Human-Video Scale and Robot Adaptation Affect Improvement?}
Tab.~\ref{tab:ablation_human_pretrain} reports results when varying the fraction of either human-video pretraining data or autonomous robot adaptation data (Sec.~\ref{sec:method}), with full data for the other stage. Human 0\% omits human-video pretraining. Adapt.~0\% omits only the initial task-agnostic adaptation phase; task-level dynamics/value fine-tuning is retained. Final success increases from 52.0\% without human pretraining to 60.0\% with half of the corpus and 85.3\% with the full corpus. Skipping initial adaptation reduces success to 65.3\%, with 68.0\% and 78.7\% at 25\% and 50\% of the adaptation data. Thus, both human-video pretraining and initial robot adaptation contribute to the final policy, even with task-level fine-tuning. This initial adaptation phase requires no teleoperated trajectories.

\paragraph{What Makes a Good Representation for Dynamics Models?} 
We ablate the choice of world-state representation in Tab.~\ref{tab:ablation_representation}. 
V8 uses point trajectories $\bm{P}_t$ only, while V9 uses visual tokens $\bm{z}_t$ only. 
V10 follows the same visual-only setting as V9, but replaces the DINO-v3 backbone with an encoder from a video generation model~\cite{wan2025}.
Both single-modality variants degrade performance: \emph{Points Only} obtains 68.0\%, and \emph{Visual Only} obtains 72.0\%, compared to 85.3\% for \textbf{Ours}. 
Replacing DINO-v3 with the video encoder (V10) lowers success from 72.0\% to 65.3\%, favoring DINO-v3 features in the visual-only setting under this training and evaluation protocol. Together, these results support combining DINO-v3 visual tokens with point trajectories for policy improvement.
\begin{table}[t]
\centering
\setlength{\tabcolsep}{4pt}
\fontsize{7}{8}\selectfont
\renewcommand{\arraystretch}{0.95}
\begin{tabularx}{\resulttablewidth}{@{}r|*{5}{Y}|Y@{}}
\toprule
\textbf{Variant}
& \textbf{\texttt{T1}}
& \textbf{\texttt{T2}}
& \textbf{\texttt{T3}}
& \textbf{\texttt{T4}}
& \textbf{\texttt{T5}}
& \textbf{Avg.} \\
\midrule
Points Only [V8]
& 60.0
& \second{66.7}
& 60.0
& \second{80.0}
& \second{73.3}
& 68.0 \\
Visual Only [V9]
& \second{73.3}
& \second{66.7}
& \second{66.7}
& 66.7
& \best{86.7}
& \second{72.0} \\
Video Latent [V10]
& 66.7
& 60.0
& \second{66.7}
& 66.7
& 66.7
& 65.3 \\
\midrule
\textbf{Ours}
& \best{86.7}
& \best{80.0}
& \best{80.0}
& \best{93.3}
& \best{86.7}
& \best{85.3} \\
\bottomrule
\end{tabularx}
\caption{Ablation of world-state representation choices, evaluated on success rate (\%).}
\label{tab:ablation_representation}
\end{table}

\section{Conclusion}
\label{sec:conclusion}
We presented Dynamics-Guided Action Correction (DGAC), a mechanism for turning failed robot rollouts into corrective supervision for policy improvement. DGAC uses successful references to guide alternative action generation and selects candidates by predicting and evaluating their outcomes. Dynamics and value models pretrained on human videos and adapted with robot experience support this process, allowing the policy to learn from failures without human-provided corrective actions.

Across seven real-world manipulation tasks on two robot embodiments, our framework improves average success from 40.0\% to 81.0\%. DGAC also improves $\pi_{0.5}$, a policy pretrained on robot data, extending the correction procedure beyond our human-pretrained policy. DGAC yields greater policy improvement than alternative ways of generating corrective supervision using the same models and rollout data.

More broadly, our findings suggest a complementary direction for robot learning. While demonstrations provide a foundation for skill acquisition, failures during deployment offer opportunities for further improvement. Combining human-video priors with embodiment-specific robot experience offers a promising path toward robots that generate their own corrective supervision and continue learning through interaction.

\paragraph{Limitations and Future Work.}
While our framework improves policies without online human corrections, several challenges remain. First, it relies on human annotations of rollout outcomes; reliable automatic outcome assessment would reduce this supervision requirement. Second, DGAC corrects short-horizon action chunks, limiting its ability to recover from failures that require longer sequences of coordinated actions. Longer-horizon prediction and hierarchical planning may help address such failures. Third, the autonomous adaptation phase relies on an affordance model and a motion planner to bring the gripper into contact with objects, so its data mainly cover simple contact interactions. Removing this dependence would extend adaptation to more complex interactions. Finally, correction depends on compatible successful references and reliable dynamics and value predictions. Expanding reference coverage and accounting for model uncertainty may extend correction to a broader range of failure states.

\acknowledgments{
This work was supported by Technical University of Munich (TUM) and the State of Bavaria through the REACT project, TUM Georg Nemetschek Institute via the SPAICR project, Munich Center for Machine Learning (MCML) and ETH Zurich. We thank Helen Oleynikova for her support during the initial phase of the project.
}

\bibliographystyle{IEEEtran}
\bibliography{root.bib}

\clearpage
\setcounter{page}{1}

\twocolumn[{\centering\LARGE\scshape Appendix\par\vspace{1.5em}}]
\pdfbookmark[1]{Appendix}{appendix}

\setcounter{section}{0}\setcounter{subsection}{0}
\renewcommand{\thesection}{A.\arabic{section}}
\renewcommand{\thefigure}{A\arabic{figure}}
\renewcommand{\thetable}{A\arabic{table}}
\renewcommand{\theequation}{A\arabic{equation}}
\renewcommand{\thepage}{A\arabic{page}}
\renewcommand{\thesubsectiondis}{\thesection.\arabic{subsection}}

\section{Supplementary Video}
We include a supplementary video on our project website showcasing an overview of our framework, along with demonstrations of various real-world robot manipulation tasks.

\section{Implementation Details}

\subsection{Human Video Dataset Processing}
We construct the human-video pretraining dataset from HOI4D~\cite{liu2022hoi4d}, Arti4D~\cite{werby2025articulated}, EgoDex~\cite{hoque2025egodex}.
For each frame, we extract semantic visual tokens using DINO-v3~\cite{simeoni2025dinov3} and obtain point-trajectory supervision with the off-the-shelf 3D point tracker TAPIP3D~\cite{tapip3d}.
Since TAPIP3D requires camera parameters and dense depth, we use the SLAM poses and camera intrinsics provided by the datasets and estimate dense depth with DepthAnything-v3~\cite{lin2025depth} for EgoDex, where dense depth annotations are unavailable.
We obtain wrist trajectories from dataset annotations when available and estimate them automatically otherwise.
For videos without wrist annotations, we first use HaMeR~\cite{pavlakos2024reconstructing} to initialize the wrist translation in the first frame.
We then sample 3D points around the initialized wrist location and propagate the wrist translation over time using TAPIP3D point flows.
Wrist orientations are estimated by HaMeR and temporally smoothed with a Savitzky--Golay filter~\cite{savitzky1964smoothing}.
Overall, this pipeline produces approximately one million samples for human-video pretraining.

\subsection{Robot Dataset Collection}
During the robot adaptation phase, the robot autonomously interacts with the scene using the human-pretrained policy. 
This procedure yields approximately 400 robot episodes in about three hours, providing task-agnostic interaction data that captures robot-specific states, contacts, action effects, and natural failures.
For each downstream task, we collect 25 teleoperated demonstrations, which are also used to initialize the task policy before self-improvement. 

\subsection{Control Strategy for Autonomous Robot  Adaptation}
Directly deploying the human-pretrained policy for robot exploration is challenging since human videos provide useful interaction priors but often lack the precision required for robot contact localization. 
We therefore collect autonomous robot interaction data using an open-loop control strategy. 
For each scene, we first use the pretrained affordance model from VidBot~\cite{chen2025vidbot} to detect plausible contact regions. 
We then select a target contact point and plan a pre-contact trajectory that moves the gripper to the corresponding region while avoiding collisions.
Once the gripper reaches the contact region, we query the human-pretrained policy to generate the post-contact interaction trajectory. 
The robot follows the predicted end-effector poses using its low-level controller, while the gripper is smoothly adjusted along the trajectory to improve execution stability. 
This strategy separates coarse contact acquisition from post-contact interaction, allowing the robot to explore semantically meaningful behaviors even when the human-pretrained policy does not provide precise contact poses. The collected rollouts contain both successful and failed interactions caused by robot-specific contacts, dynamics, and scene configurations. We use these autonomous rollouts to adapt the dynamics and value models before task-level self-improvement.

\subsection{Baseline Implementation Details}

\paragraph{SWIM.}
Since SWIM~\cite{mendonca2023SWIM} is not open-sourced, we re-implement the baseline ourselves. 
We use the same architectural configurations as ours for both the affordance and dynamics models, but modify the affordance head to predict SWIM's structured action space. 
SWIM is originally demonstrated on relatively short-horizon goal-conditioned manipulation tasks, such as object picking or drawer opening. 
To extend it to our long-horizon tasks, we use the same goal-image retrieval strategy as ours and rank candidate corrections by the distance between the predicted future feature and the retrieved goal feature.

\paragraph{LPB.}
LPB~\cite{sun2025lpb} was originally formulated as a plug-and-play module for diffusion policies~\cite{chi2023diffusionpolicy}, where differentiable guidance is used to steer action generation.
For a fair comparison, we adapt LPB to our flow-matching policy by using the gradient of the latent distance-to-goal objective to guide the inferred velocity field during action generation. 
As LPB's nearest-neighbor search scales poorly with the size of the teleoperation dataset, we use our value model to accelerate goal retrieval. 

\paragraph{AWR.}
We follow AWR~\cite{peng2019AWR} to update the policy with advantage-weighted behavioral cloning. We convert the estimated advantage into a sample weight using its proposed exponential form. The resulting weights are clipped for numerical stability and used to weight the behavior-cloning objective during policy training.

\paragraph{RECAP.}
RECAP~\cite{intelligence2025pi} is closely related to CFGRL~\cite{frans2025diffusion}: both convert advantage estimates into conditioning embeddings and train the policy under different conditioning modes. Following RECAP, we estimate the advantage for all states in the collected dataset and use the 70th percentile of the empirical advantage distribution as the cutoff. Samples above this cutoff are trained with the optimal embedding, while the remaining samples are trained with the non-optimal embedding. For a fair comparison in our autonomous self-improvement setting, we remove human-in-the-loop corrections from RECAP.

\paragraph{RISE.}
RISE~\cite{yang2026rise} extends RECAP by introducing a video generator to synthesize additional rollouts for policy co-training.
We initialize from their provided video model checkpoints and fine-tune the model using our collected robot rollout data. It shares the same action space as our dynamics model, which is represented using the wrist 6-DoF pose and closure parameter.

\subsection{Network Architecture}
\paragraph{Policy Model.}
Our policy model follows a transformer-based action-chunking architecture similar to Diffusion Policy~\cite{chi2023diffusionpolicy}. 
We implement it as an encoder-decoder transformer that predicts future end-effector action chunks ${\bm{a}}_{t:t+H-1}$ under the flow-matching objective. 
The model conditions on the language instruction, visual history, and wrist-state history $\bm{s}_{t-H'+1:t}$. 
These inputs are projected into a shared latent space and encoded as conditioning tokens, together with the flow-matching timestep and advantage embeddings when applicable. 
The decoder takes noisy action tokens with learnable positional embeddings and predicts the clean future action chunk using causal self-attention. 
The policy outputs actions in the world frame with a chunk size of $H=30$. 
We use 16 transformer decoder layers, 12 attention heads, and a latent dimension of 384.
In the main text, we denote the auxiliary policy conditions as $\bm{m}_t^\pi$, which include language tokens and camera-aware ray maps~\cite{wang2025continuous}. 
Language tokens are extracted using a pretrained text encoder~\cite{raffel2020exploring}. 
For visual history, we extract DINO-v3 visual tokens~\cite{simeoni2025dinov3} from head-camera observations and concatenate them with the corresponding ray maps. 
During robot deployment, we additionally include DINO-v3 visual tokens from the gripper camera, together with the corresponding gripper-camera ray maps. 
Camera extrinsics are represented as relative transformations between the current camera pose and historical camera poses. 
All head-camera images used during human pretraining and robot deployment are center-cropped and resized to $224\times224$, while gripper-camera images are resized to $240\times320$.

To accommodate bi-manual demonstrations in human videos, we represent each wrist state and action using two 6-DoF wrist poses, each paired with a scalar closure value. Rotations are parameterized using the continuous 6D representation from~\cite{zhou2019continuity}, resulting in a 20-dimensional action representation: $2 \times (3$ translation $+ 6$ rotation $+ 1$ closure$)$. During robot deployment, we preserve this two-hand format for compatibility. For single-arm robots, we duplicate the robot action into the unused hand slot only as a placeholder; this placeholder is ignored by the controller, and only the executable robot action is applied. For human-video pretraining, we set the history horizon $H'$ to 15; for robot policy training, we set it to 1.

\paragraph{Dynamics Model.}
The dynamics model follows the CDiT transformer architecture~\cite{bar2024navigationworldmodels,peebles2023scalable}. 
It predicts the future state representation $\hat{\bm{o}}_{t+H}$ conditioned on the history state $\bm{o}_{t-H'+1:t}$, a candidate action chunk $\bm{a}_{t:t+H-1}$, and auxiliary context tokens $\bm{m}_t^f$.
In our implementation, $\bm{m}_t^f$ includes language tokens extracted by T5~\cite{raffel2020exploring} and camera ray maps, which are concatenated with the visual tokens for conditioning. 
Under the flow-matching objective, the noisy future-state sample is patchified and linearly projected into latent tokens. 
The historical state maps are also patchified and projected into latent tokens, while context tokens and action tokens are embedded with separate MLPs. 
The history, context, and action tokens are then concatenated to form the conditioning sequence. 
We add learnable absolute positional embeddings to the noisy future-state patches, history patches, context tokens, and action tokens.
Each transformer layer performs self-attention over the noisy future-state tokens, followed by cross-attention to the conditioning sequence and an MLP. 
Following DiT-style conditioning~\cite{peebles2023scalable}, the flow-matching timestep is encoded as a global conditioning vector and used to modulate the self-attention, cross-attention, and MLP layers through adaptive LayerNorm. 
The model predicts the clean future state $\hat{\bm{o}}_{t+H}$, represented by future visual tokens $\hat{\bm{z}}_{t+H}$ and point trajectories $\hat{\bm{P}}_{t+H}$.
We use 16 transformer layers, 12 attention heads, and a latent dimension of 384. 
All camera images used during human pretraining and robot deployment are center-cropped and resized to $224\times224$. We set the history horizon $H'$ of the dynamics model to 15 for both human-video pretraining and robot training.

\paragraph{Value Model.}
The value model uses a transformer-based conditioning architecture to predict a scalar value for the current state. 
It takes the state representation $\bm{o}_t$ as input and conditions on auxiliary information $\bm{m}_t^V$, including language tokens extracted by T5~\cite{raffel2020exploring} and the current wrist state $\bm{s}_t$. 
We use a learnable value token as the decoder query to attend to the encoded state and auxiliary conditioning tokens. 
The output feature corresponding to this value token is passed through a prediction head to produce the scalar value estimate $\hat{V}_t$. 
This value measures task progress and is used to rank candidate corrective actions during DGAC. 
We implement the value model with 8 transformer layers, 8 attention heads, and a latent dimension of 384. 
All camera images used during human pretraining and robot deployment are center-cropped and resized to $224\times224$.

\subsection{Training Details}
\paragraph{Human-Video Pretraining.}
For human-video pretraining, we train the action policy, dynamics model, and value model jointly for 250k iterations with a batch size of 20 on 4 GPUs. 
We use AdamW~\cite{loshchilov2017decoupled} with a learning rate of $1\times10^{-4}$. 
Since the dynamics model predicts both point flows and visual tokens, we use separate loss terms for the two prediction targets and weight the visual dynamics loss by 100.

\paragraph{Robot Adaptation and Policy Initialization.}
For robot adaptation, we train the dynamics and value models for 80k iterations with a batch size of 20 on 4 GPUs. 
We use the same AdamW optimizer with a learning rate of $1\times10^{-4}$. 
To initialize the task policy, we train it on the task-specific teleoperation data for 30k iterations with a batch size of 32 on 2 GPUs.

\paragraph{Per-Task Fine-Tuning during Self-Improvement.}
Starting from the adapted dynamics and value models, we further fine-tune them during each self-improvement iteration for 8k iterations with a batch size of 12 on 2 GPUs. 
We again use AdamW with a learning rate of $1\times10^{-4}$.
For value-model training, we weight the temporal-difference loss $\mathcal{L}_{\text{TD}}$ by a factor of 10.
The policy is updated in each self-improvement loop using the same number of training iterations as in the policy initialization stage and uses the same weight initialization. 
For advantage-conditioned policy extraction, we compute the advantage threshold $\kappa$ from the advantage distribution and use its 70th percentile as the cutoff. We run two self-improvement iterations for all methods and collect 20 robot episodes in each iteration.

\section{Manipulation Task Design Details}
We evaluate policy improvement on seven real-world manipulation tasks across two robots, covering diverse long-horizon household interactions. The tasks involve reaching, grasping, articulated-object manipulation, placement, and task-conditioned sequencing, with randomized object appearances and scene layouts. Their multi-stage structure and contact-rich dynamics induce diverse failure modes, providing a challenging testbed for evaluating whether human-video priors and DGAC-based self-improvement can turn robot failures into useful supervision.

\subsection{Manipulation Tasks for Hello Robot Stretch 3}
\paragraph{Manipulation Task: Socks.}
\begin{itemize}
    \item \textbf{Initial state:} Two socks with randomized colors and positions are placed on the tabletop. The drawer is initially open. A basket is placed on either the left or right side of the table and may be empty or contain one sock.
    
    \item \textbf{Subgoals:} 
    (a) Reach toward one sock and pick it up;
    (b) Place it into the drawer;
    (c) Reach toward the remaining sock and pick it up;
    (d) Place it into the drawer;
    (e) Close the drawer.
    
    \item \textbf{Success criterion:} The task succeeds if both socks are placed inside the drawer and the drawer is fully closed.
\end{itemize}

\paragraph{Manipulation Task: Kitchen.}
\begin{itemize}
    \item \textbf{Initial state:} A rice cooker is positioned on the left side, and a pan is placed on either the left or right burner of the stove. A randomly selected vegetable is placed either in the pan or in the sink. The rice cooker is initially closed.

    \item \textbf{Subgoals:}
    (a) Press the button to open the rice cooker lid.
    (b) Reach toward the vegetable.
    (c) Grasp the vegetable.
    (d) Transport the vegetable to the rice cooker.
    (e) Place the vegetable into the rice cooker.
    (f) Close the rice cooker lid.

    \item \textbf{Success criterion:} The task is successful if the vegetable is placed inside the rice cooker and the rice cooker lid is fully closed.
\end{itemize}

\paragraph{Manipulation Task: Microwave.}
\begin{itemize}
    \item \textbf{Initial state:} A microwave with its door initially closed is placed on the tabletop. A drink with a randomized shape and color is placed inside the microwave, and a blue plate is randomly placed in the center-left region of the counter.
    \item \textbf{Subgoals:}
    (a) Open the microwave door.
    (b) Reach toward the drink inside the microwave.
    (c) Take the drink out of the microwave.
    (d) Place the drink onto the blue plate.

    \item \textbf{Success criterion:} The task is successful if the drink is removed from the microwave and placed upright on the blue plate.
\end{itemize}

\paragraph{Manipulation Task: Table.}
\begin{itemize}
    \item \textbf{Initial state:} A tabletop cleaning setup is placed in front of the robot. A sponge is placed on either the left or right side of the table, and a visible stain is located near the center of the tabletop.

    \item \textbf{Subgoals:}
    (a) Reach toward the sponge.
    (b) Grasp the sponge from the tabletop.
    (c) Move the sponge toward the stained region.
    (d) Wipe the tabletop until the stain is removed.
    (e) Place the sponge back on the table.

    \item \textbf{Success criterion:} The task is successful if the visible stain is sufficiently removed from the tabletop and the sponge is placed back on the table after cleaning.
\end{itemize}

\paragraph{Manipulation Task: Basketball.}
\begin{itemize}
    \item \textbf{Initial state:} A mini basketball hoop is mounted at the back-left side of the workspace. A basketball is placed on a small sofa with a randomized initial position.

    \item \textbf{Subgoals:}
    (a) Reach toward the basketball.
    (b) Grasp the basketball from the sofa.
    (c) Lift the basketball and move it toward the hoop.
    (d) Drop the basketball through the hoop.

    \item \textbf{Success criterion:} The task is successful if the basketball passes through the hoop and lands in the box.
\end{itemize}

\subsection{Manipulation Tasks for the Franka Panda}
\paragraph{Manipulation Task: Box.}
\begin{itemize}
    \item \textbf{Initial state:} A target object and a box are randomly placed on the workspace.
    
    \item \textbf{Subgoals:} 
    (a) Reach toward the target object and grasp it;
    (b) Transport the object to the box;
    (c) Place the object into the box.
    
    \item \textbf{Success criterion:} The task is successful if the target object is placed inside the box.
\end{itemize}

\paragraph{Manipulation Task: Sweep.}
\begin{itemize}
    \item \textbf{Initial state:} A brush and a dustpan are placed on the workspace with randomized positions, either to the left or right of the robot. 
    Several 3D-printed objects with randomized shapes and colors are placed between them.

    \item \textbf{Subgoals:}
    (a) Reach toward the brush and grasp it;
    (b) Keep swiping until all objects are in the dustpan;
    (c) Place the brush down after all objects have been swept into the dustpan.
    
    \item \textbf{Success criterion:} The task is successful if all objects are placed inside the dustpan and the robot places the brush down afterward. 
    The robot should neither continue sweeping after all objects are inside the dustpan nor place the brush down before completing the sweep.
\end{itemize}

\end{document}